\documentclass[letterpaper]{article} 
\usepackage{aaai24}  
\usepackage{times}  
\usepackage{helvet}  
\usepackage{courier}  
\usepackage[hyphens]{url}  
\usepackage{graphicx} 
\def\UrlFont{\rm}  
\usepackage{natbib}  
\usepackage{caption} 
\usepackage{algorithm}
\usepackage{algorithmic}

\usepackage{newfloat}
\usepackage{listings}
\DeclareCaptionStyle{ruled}{labelfont=normalfont,labelsep=colon,strut=off} 
\floatstyle{ruled}
\newfloat{listing}{tb}{lst}{}
\floatname{listing}{Listing}
\usepackage[caption=false]{subfig}

\title{On Discprecncies between Perturbation Evaluations of \\Graph Neural Network Attributions}
\author{
    Razieh Rezaei\textsuperscript{\rm 1,4}\equalcontrib,
    Alireza Dizaji\textsuperscript{\rm 2}\equalcontrib,
    Ashkan Khakzar\textsuperscript{\rm 1},\\
    Anees Kazi\textsuperscript{\rm 1,3},
    Nassir Navab\textsuperscript{\rm 1},
    Daniel Rueckert\textsuperscript{\rm 1}
}
\affiliations{

    Technical University of Munich\textsuperscript{\rm 1},
    Sharif University of Technology\textsuperscript{\rm 2}, \\
    Harvard Medical School\textsuperscript{\rm 3},
    Hospital of the Ludwig-Maximilians-University (LMU) Munich\textsuperscript{\rm 4}

}

\usepackage{bibentry}
\usepackage{amsmath}

\begin{document}
\maketitle
\begin{abstract}
    Neural networks are increasingly finding their way into the realm of graphs and modeling relationships between features. Concurrently graph neural network explanation approaches are being invented to uncover relationships between the nodes of the graphs. However, there is a disparity between the existing attribution methods, and it is unclear which attribution to trust. Therefore research has introduced evaluation experiments that assess them from different perspectives. In this work, we assess attribution methods from a perspective not previously explored in the graph domain: retraining. The core idea is to retrain the network on important (or not important) relationships as identified by the attributions and evaluate how networks can generalize based on these relationships. We reformulate the retraining framework to sidestep issues lurking in the previous formulation and propose guidelines for correct analysis. We run our analysis on four state-of-the-art GNN attribution methods and five synthetic and real-world graph classification datasets. The analysis reveals that attributions perform variably depending on the dataset and the network. Most importantly, we observe that the famous GNNExplainer performs similarly to an arbitrary designation of edge importance. The study concludes that the retraining evaluation cannot be used as a generalized benchmark and recommends it as a toolset to evaluate attributions on a specifically addressed network, dataset, and sparsity. Our code is publically available.\footnote{
    https://github.com/alirezadizaji/GraphROAR
    }
\end{abstract}

\section{Introduction}

\begin{figure}[!ht]
    \centering
  
    \includegraphics[width=0.80\columnwidth]{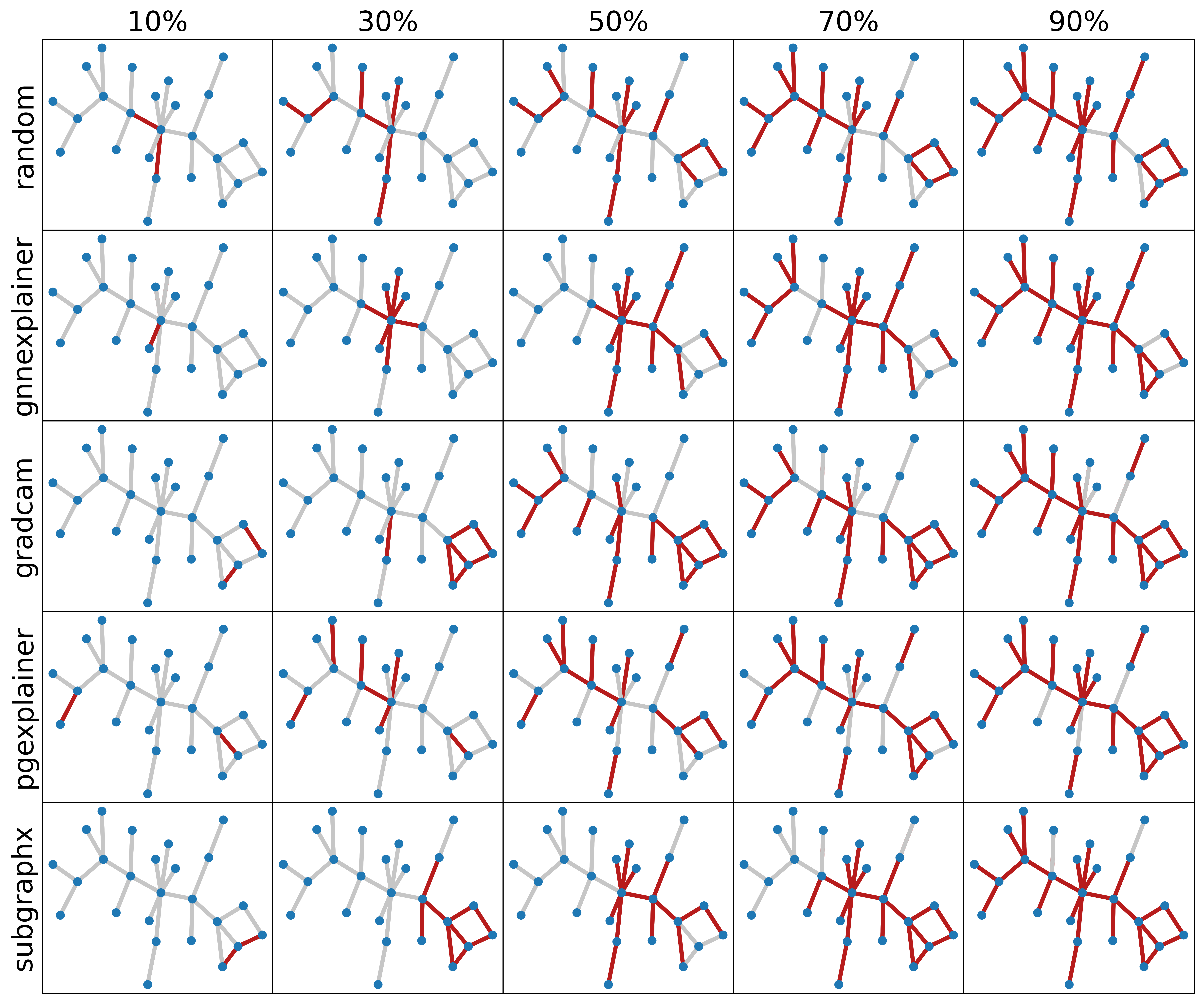}
    \caption{\textbf{Attribution Disagreement:} Important edges (top 10, 30, 50, 70, 90 $\%$) as identified by different methods (each row) are different (BA2Motif dataset)}
   \label{fig1:teaser}
   
\end{figure}

Attribution methods for Graph Neural Networks (GNNs) are coming into the spotlight to alleviate the issues arising from the black-box behavior of GNNs \cite{gnn_interpretability_Survey, gnnExp, yuan2020xgnn, pgExp, subGraphX, sanchez2020evaluating, agarwal2022probing}.
The attribution helps establish trust, debug failure modes, and extract insights regarding predictive node relationships and features within the data. Feature attribution research has been ongoing before finding its way to graph domain\cite{trust, cv1, cv2, cv3, cv4, zhang2021fine, gradCam_Vision,Khakzar_2021_CVPR}. Some of these methods such as GradCAM \cite{gradCam} are directly transferred to the graph domain. Some are directly proposed for graphs considering the properties of graphs: For instance, optimization-based methods such as GNNExplainer \cite{gnnExp}, PGExplainer \cite{pgExp}, and search algorithms (SubGraphX \cite{subGraphX}) to identify the important subset of nodes and edges for the classification GNN model.\

Despite the progress in explainability research, the problem remains unsolved, and there is a disagreement between different attribution method outputs \cite{krishna2022disagreement, khakzar2022explanations}. It is not clear which attribution to trust. The same phenomenon exists in graph attribution (See Figure~\ref{fig1:teaser}). Some evaluation methodologies exist to check the sanity of explanations \cite{samek2016evaluating, ROAR, adebayo2018sanity, sanchez2020evaluating, agarwal2022probing, zhang2023attributionlab}. Some approaches solely rely on human judgment to evaluate interpretability \cite{pgExp_binary, pgExp_reparTrick, gnnExp, yuan2020xgnn}. These approaches compare the attribution with ground truth features. For instance, in the image domain, several metrics (pointing game \cite{zhang2018top}, EHR \cite{zhang2021fine}, IoU \cite{gradCam_Vision}) compare attribution (saliency maps) against bounding box annotations. There are counterpart approaches in the graph domain that compare attributions with human-selected subgraphs \cite{pgExp_binary, pgExp_reparTrick, gnnExp, yuan2020xgnn}. However, these ground truths are provided by humans, and there is no guarantee that the model would use the same features as humans use \cite{samek2016evaluating, ROAR, adebayo2018sanity}. Such metric can lead to false explanations as the model may choose a different combination of features than a human expert to achieve the same accuracy. In this work, we investigate the significance of the edges in the graph. 

But, how do we know if an edge in the graph is important for the GNN? Remove it and observe how the output changes. This notion is the motivation behind the perturbation-based analysis of attributions. Originally \cite{samek2016evaluating} proposed a framework based on perturbation, and this notion was later incarnated in the graph domain as the fidelity vs. sparsity score \cite{subGraphX,li2022explainability}. In simple terms, observing the GNN's output (through fidelity) for various levels of sparsity as we keep removing edges in the order of their importance. Despite being insightful, there is always the uncertainty that the resulting output change from removing edges is due to the new perturbed input being out of distribution since the model has never seen such graph subsets during training.

The idea of retraining on attributions is proposed in \cite{ROAR} emerged to overcome this issue. However, it never found its way to the graph domain, and the evaluations are still based on variants of fidelity vs. sparsity. The retraining evaluation unveiled novel insights regarding the behavior of several attributions. For instance, it is shown that the features identified as important network's gradient are not generalizable features, although removing these features can affect the output significantly. 

In this work, we first adapt the evaluation by retraining strategy to the graph domain. Moreover, we discuss issues lurking in the original formulation of the retraining framework and provide a different approach to side-step these issues. We outline how to reliably interpret the retraining evaluation results. We demonstrate the evaluation pipeline on four popular and renowned explanation methods: GradCAM \cite{gradCam}, GNNExplainer \cite{gnnExp}, PGExplainer \cite{pgExp}, and SubgraphX \cite{subGraphX}. We also add a random explainer (random edge importance assignment) to evaluate how the explanations performance deviates from a random assignment, and surprisingly, GNNExplainer (the most renowned method) never does. We leverage five datasets: two synthetic datasets (BA2Motifs \cite{gnnExp}, BA3Motifs created by us), two biology datasets (MUTAG \cite{debnath1991structure}, ENZYME \cite{schomburg2004brenda}), and a social-networking dataset (REDDIT-BINARY \cite{reddit}). We show how explainers behave differently in different datasets and networks (GIN \cite{GIN} and GCN \cite{GCN}). By demonstrating the variability in performance, we show that retraining evaluation is required whenever an attribution is used. Our contributions can be summarized as follows:

\begin{itemize}
\item We innovate by adapting the concept of retraining, previously applied in the vision domain, and reformulating it into the realm of graph data and GNNs.
\item Through extensive experiments across five datasets, we evaluate the performance of four renowned graph attribution methods and also a random setting.
\item We propose a guidebook outlining a procedural approach for interpreting attribution results and comparing different methods.
\item Our results challenge previous claims, demonstrating that no single model exhibits consistent superiority across all experiments. Instead, we show calling a model the best-performing depends on specific conditions which we later elaborate on.
\end{itemize}

\section{Background and Setup}
In this section, we introduce the notations concerning graph data, classification GNN, the employed attribution methods, and our retraining strategy. Followingly, we provide a concise overview of the background regarding the functioning of the four used attribution methods. For a more detailed elaboration on these methods, please refer to the supplementary materials.

\paragraph{Notation:}
We represent a graph as $G=\left \{ V,A \right \}$ where $V=\left \{ v_1, v_2,..., v_m, ..., v_M \right \}$ is set of $M$ nodes and $A \in R^{M\times M}$ the adjacency matrix. Each element of $A$ i.e. $a_{ij}\in  \left \{ 1,0 \right \}$ based on the presence or absence of the edge, respectively between $v_i$ and $v_j$. $V$ is also associated with a feature matrix $X \in R ^{M\times d}$ consisting of d-dimensional node features. The task at hand is to classify each graph $G$ to a class from $y \in \left \{ 1,2,...,N \right \}$. In general, we represent GNN model as $f(.)$. Below we explain the general framework of GNN attribution methods.

\paragraph{Attribution:}
Given a graph $G$, and node feature matrix $X$, a GNN's prediction is $\hat{y} = f(G, X)$ getting compared with $y$ as the graph's true label. An attribution method tries to find a subset of nodes $V$ in the original graph $G_o$ which is most important for the classification task. In other words, it generates an attribution for prediction $\hat{y} = f(G_s, X_s)$ with $G_s$ as a subset of size $m$ from the $G_o$ with $m \leq M$. The subset only sparsely includes the nodes in the original graph and thus $G_s \subseteq G_o$. In some literature, sparsity is introduced as the percentage of edges removed from $G_{o}$ after the attribution is applied. For simplicity, here we define sparsity as the percentage of \textit{edges kept} from the $G_{o}$ as
$sparsity = \frac{100*|E(G_{s})|}{|E(G_{o})|}$ with $E(G)$ indicating a set of edges of the graph. The training of a network, the GNN model, on the attributed subgraph $G_s$ with hyperparameters of $\theta$ can be specified as $f(G_s, \theta)$, while later as we elaborate on the retraining concept we use ${f}'(G_s, {\theta}')$ to show the network has been trained again on the new training set. The accuracy is defined as $N_{\text{correct}} = \sum_{i=1}^{N} I(\hat{y}_i = y_i)$.

\subsection{Attributions under Evaluation}

\textbf{GNNExplainer\cite{gnnExp}}.
In this paper, the model learns a feature mask proportional to the importance of the features. The importance of the input features is evaluated based on model's sensitivity w.r.t its presence and absence. Therefore GNNexplainer gives more weight to the features which affects models behavior. For backpropagating the gradients in mutual information maximization equation, reparameterization trick was leveraged.

\begin{equation}
\label{eq:MI}
\max_{G_s, F} MI(Y,(G_s, F)) = H(Y) - H(Y|G=G_s,X=X_{S}^{F})
\end{equation}

\textbf{PGExplainer \cite{pgExp}}. In this method, the objective is again to maximize the mutual information $MI(\cdot ,\cdot )$ between the GNN's prediction when $G$ and $G_{s}$ are the input to the model. In this equation $H\left ( \cdot  \right )$ stands for entropy. This can be seen as a minimization problem since the term $H(Y)$ in the equation \ref{eq:MI} is fixed, and as a result, the objective turns into minimizing the conditional entropy $H(Y_s|G=G_s)$. Two relaxation assumptions have been used for this model: i) $G_{s}$ is a gilbert graph with independently selected edges from $G$, and ii) probabilities of an edge $(e_{ij})$ from the original graph existing in the explanatory graph $P(e_{ij})$ is approximated with Bernoulli distribution $e_{ij} \sim Bern\theta_{i,j}$.
Also, reparameterization trick \cite{pgExp_reparTrick} is leveraged to optimize the objective function with gradient-based methods. Finally, binary concrete distribution was utilized to approximate the sampling process from the Bernoulli distribution and forming $\hat{G}_{s} \approx G_{s}$. Considering several subsets and classes, Monte Carlo is used to aproximately oprimize the objective function with $K$ as the total number of sampled graph, and $C$ as the number of labels, and $\hat{G}_s^{(k)}$ as the k-th sampled graph:
\begin{align}
&\min_{\Omega } \; E_{\epsilon \sim Uniform(0,1)} H(Y_{0},\hat{Y}_{s}) \; \approx \min_{\Psi } - \dfrac{1}{K} \sum^K_{k=1}\sum^C_{c=1} \nonumber \\
& P_{\Phi}(Y=c|G=^{(i)})\log P_{\Phi}(Y=c|G=\hat{G}_{s}^{(i,k)})
\end{align}

\textbf{GradCAM \cite{gradCam}}.
Let the output of $j^{th}$ layer be $\hat{X}^{j+1}$ and the gradient be $g_{j}\in R^{N\times C_j}$ where, $C_j$ is the number of neurons in the $j^{th}$ layer. GradCAM proposes to compute feature weights signifying the importance of the feature $\alpha_k$ as: 
$\alpha_k = \sum_{i\in C_j} \hat{X}^{j+1}_{ki}\ast O_j$ where, 
$O_j= \sum_{i=0}^{N-1}g_{j}(i)$ and $i$ denotes the neurons. Inspired from GradCAM and Dive into graphs (DiG)\cite{liu2021dig}, we rank the edges $e_{l,m}$ in the input graph $G$ by applying a normalization of the gradients throughout the network and adding the activation function to that. Here $l$ and $m$ denote $l^{th}$ and $m^{th}$ node. The rank of $e_{lm}$ is denoted by $\beta_lm$ and can be mathematically formulated as: $\beta_{lm} = \frac{\alpha_l+\alpha_m}{2} $.


\textbf{SubgraphX \cite{subGraphX}}.
This method uses the concept in game theory to gain different graph structures as players. To explore the important subgraph, Monte Carlo Search Tree has been used. In this search tree, root is the input graph, and each of other nodes corresponds to a connected subgraph. To reduce the search space MCTS is leveraged by selecting a path from a root to a leaf node with highest reward. $\displaystyle G^* = \operatorname*{argmax}_{\left|\displaystyle G_s^i\right| \leq N_{min}} Score(f(.), \displaystyle G, \displaystyle G_s^i)$ is a scoring function for evaluating the importance of a subgraph $G_s^i$ with the upper bound $N_{min}$ for its size given the trained GNNs $f(.)$ and the original input graph $G$. After several iterations of search, the subgraph with the highest score from the leaves is identified as the explanation. For both MCTS and selection of explanation, Shapley value \cite{subGraphX_shapley_1,subGraphX_shapley_2} from the cooperative game theory was use as the score function. Set of the players is defined as $P=\left \{G_s^i, v_k, ..., v_r \right \}$ where the subgraph $G_s^i$ is one player with $k$ nodes and $r (r \leq m-k)$ is the number of its L-hop neighboring nodes. To have an even more efficient computing, Monte Carlo sampling is applied to calculate Shapley values with considering a $S_i = P\setminus\left\{G_s^i\right\}$ as a coalition set of nodes. Finally, the approximation of Shapley values $\phi(G_s^i)$ with $T$ total number of sampling steps and $f(.)$ as the trained GNN network, can be written as: $\phi (G_s^i) = \frac{1}{T}\sum_{t=1}^{T (f(S_{i}\cup{{G_s^i}})-f(S_{i}))}$

\section{Attribution Evaluation}
The core idea behind evaluation by retraining is simple:
first \textit{Retrain} the network on the \textit{subgraph} (a subset of nodes and edges of the original graph) determined by the attribution method, and then check the network’s generalization ability. We can do this in two variations which give complementary insights:

\textbf{RoMie: Retrain On the Most Important Edges}.
For each input graph in the dataset, an attributed graph subset (\textit{subgraph}) is extracted. The subgraphs form a new training set consisting of edges identified as important by the employed attribution method. Then the network is trained again from a random initialization state on this new training set. Finally, the network generalization ability on the test set is evaluated. If the network exhibits satisfactory generalization performance by exclusively relying on the attributed edges during training, it can be inferred that selected edges possess highly predictive value for that particular network under investigation. Additionally, RoMie must be applied to varying degrees of sparsity in graphs in order to have a fair valuation. For a given sparsity level of $N\%$, the most important edges, or the top $N\%$ determined by the attribution method, are chosen to form a subgraph. By altering the percentage N, an extensive and fair analysis of the attribution method's performance across different sparsity levels can be conducted. Of particular note is that the original test set will be used for testing purposes. We will provide further explanation for this decision in a separate subsection later.

\textbf{RoLie: Retrain On the Least Important Edges}.
For each input graph in the dataset, a non-attributed graph subset (\textit{subgraph}) is extracted. In this setting, the training set is formed of subgraphs that include edges with the Least Importance; meaning, for a given sparsity level of $N\%$, the least important edges, or the lowest $N\%$ determined by the attribution method, are chosen to form a subgraph. The network is then trained on these subgraphs. In a similar fashion, the test set will be the original unperturbed set. Further elaboration on this, will be provided later on. Finally, the performance of the attribution method is again studied across different levels of sparsity for a fair comparison.

\begin{figure}[!ht]
    \centering

    \includegraphics[width=0.70\columnwidth]
    {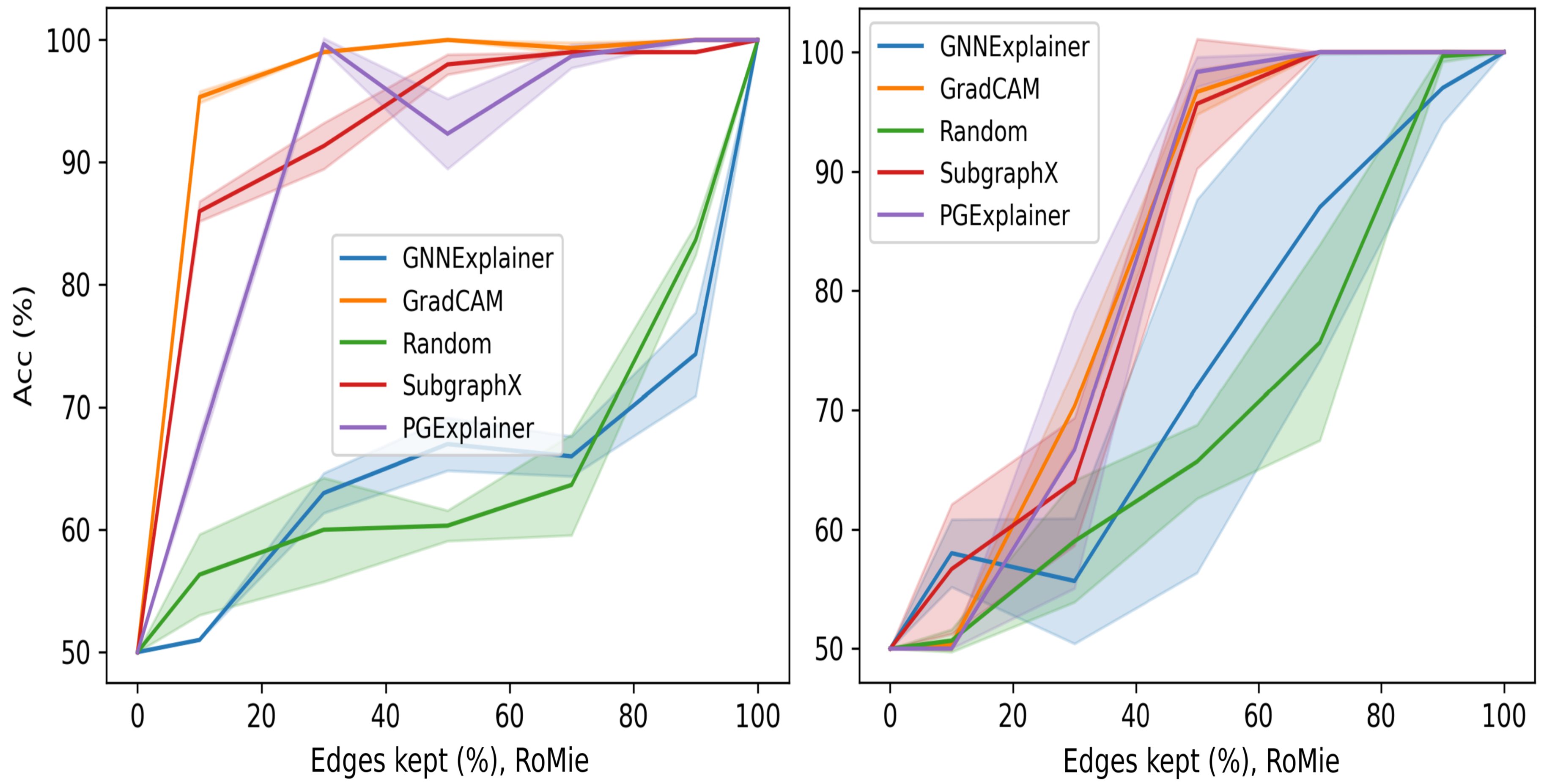}
    \caption{
    \textbf{unperturbed} (left) \textbf{vs perturbed} (right) \textbf{test sets}: There is a sudden rise in accuracy in the unperturbed version as the network performs well even in very small percentages of sparsity, on BA2Motif, for GCN.
    }
   \label{fig2:testset}

\end{figure}

\textbf{How to correctly interpret RoMie and RoLie?}
The two variations give complementary views: In a nutshell, RoMie tells us if an attribution method \textit{succeeds} in identifying sufficiently predictive edges, while RoLie tells us if an attribution method \textit{misses} to mark all predictive edges. In both scenarios, the attribution checks whether the remaining edges are generalizable by the network. Note that we cannot claim anything for certain in case the network does not generalize using the remaining edges (as opposed to what is claimed in [7]) because the inability to generalize might stem from factors such as optimization, OOD effects, network properties, and other unforeseen elements. However, if the network generalizes, we can confidently state the remaining edges were predictive. Therefore, we need to exercise caution in interpreting RoMie and RoLie, and we propose the following rulebook for each approach:

\textbf{Interpreting RoMie:}
for a sparsity under consideration, when \textit{Retraining} on the \textit{Most} important edges selected by the attribution results in the original accuracy, that means those edges have a high contribution to network predictivity. If the accuracy is high specifically for smaller subgraphs, when the sparsity level is low, we can say an attribution method is a top performer in terms of precisely identifying these predictive edges at such low sparsity. Clearly, the accuracy of a desirable attribution method rises to the original more quickly as we add more edges.

\textbf{Interpreting RoLie:}
for a sparsity under consideration, if we get the original accuracy even though the \textit{Retraining} is done on the \textit{Least} important edges instead of the most important ones, that is a sign of the subgraph's undesirable predictive power. It means that the attribution method is missing to select all predictive edges and some still remain in the subgraph of the least importance. In the specific case of low sparsity, if the network still generalizes, we can say the attribution method is not desirable. Therefore in the RoLie plots, we expect to observe a start from a low point followed by a smooth rise in accuracy for a top-performing method.

\begin{figure}[!ht]
    \centering

    \subfloat[BA3Motif]{%
      \includegraphics[clip,width=0.7\columnwidth]{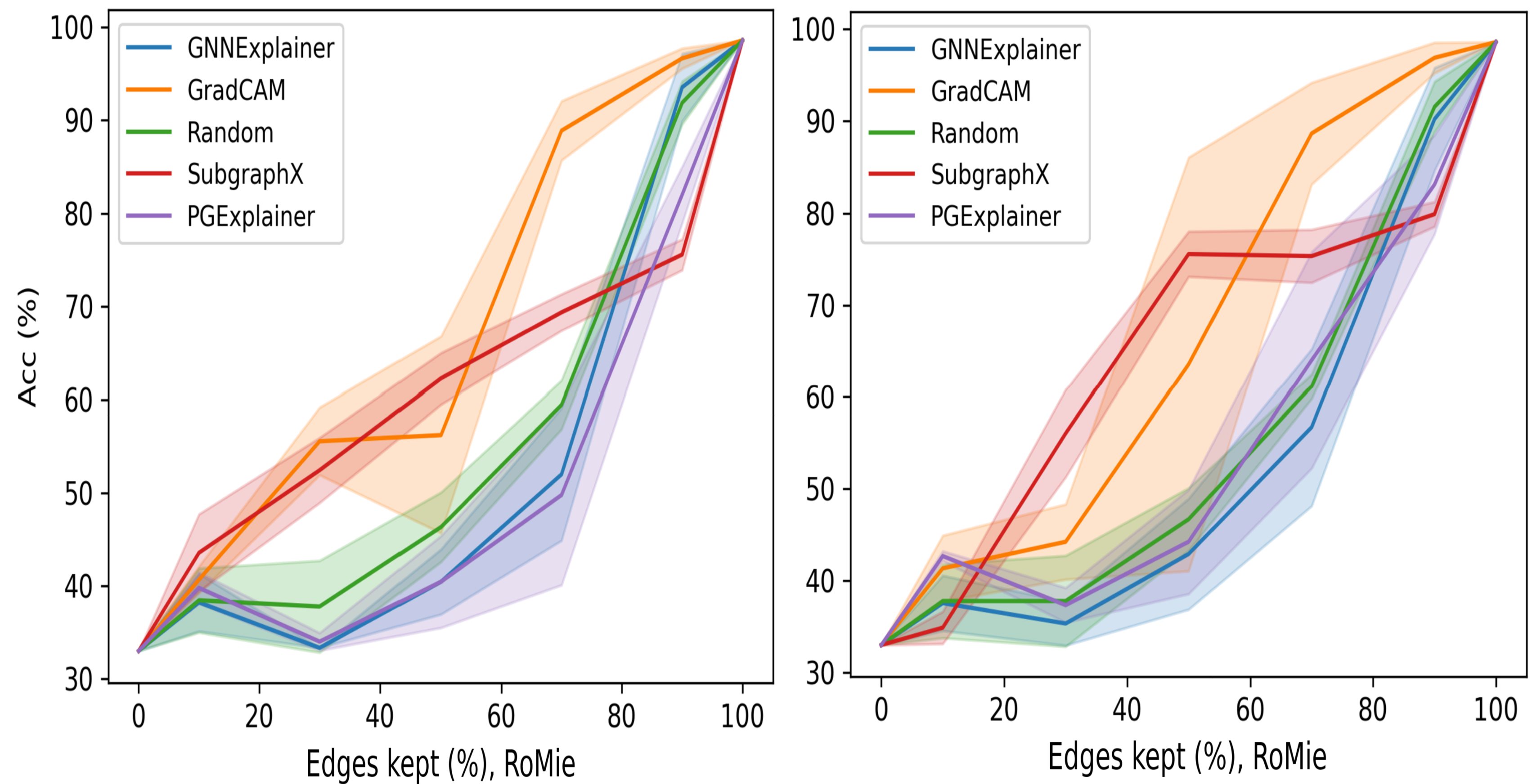}%
    }%

    \subfloat[Reddit]{%
      \includegraphics[clip,width=0.7\columnwidth]{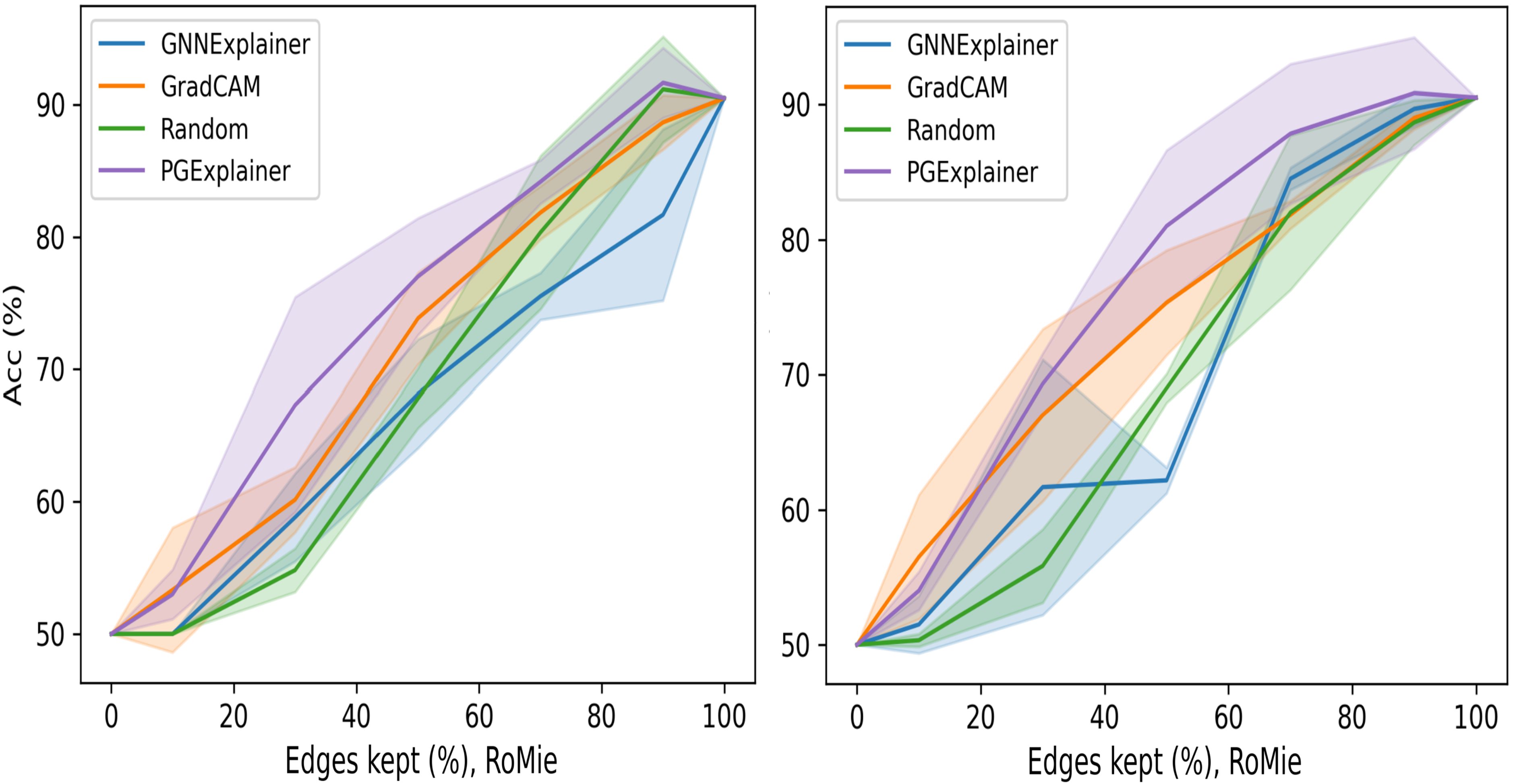}%
    }

    \caption{\textbf{Elimination of Isolated Nodes:}
    {\textbf{with}} (left), and {\textbf{without}} (right) isolated nodes.
    No significant change in performance of attribution methods is observed while eliminating isolated nodes in synthetic datasets.
    {\textbf{a}} on BA3Motif, and {\textbf{b}} on Reddit, both for GCN
    }
   \label{fig3:nodeEliSynth}
   
\end{figure}

\subsection{Perturbed vs. Unperturbed Test Set}
Earlier, we mentioned using the unperturbed test set for evaluation of RoMie and RoLie. The motivation behind using the original, unperturbed test set is two-fold. \textbf{Firstly}, to ensure a fair comparison of network accuracy across various levels of sparsity, it's essential to maintain consistency in the test set used across all sparsity conditions. Comparison of accuracies loses its reliability when the test sets differ in each evaluation due to varying perturbations. As a result, the impact of adding or removing important edges into the subgraph in accuracy plots will not be readily trackable (Figure~\ref{fig2:testset}).
\textbf{Secondly}, based on our observations in Figure~\ref{fig6:RoMieRoLie} (b) and (c), some specific structural patterns (cycle or house for BA2Motif) might form during perturbation in graph datasets. These patterns could potentially introduce a classification bias. The intention behind attribution methods is to support the network to make classification decisions based on the existing features of nodes and edges within a subgraph. The patterns, however, might serve as favorable but dishonest cues for the network to correctly classify without genuinely considering those features. To avoid these patterns in the test set, we once again recommend sticking to the original test set. 

\textbf{How about Out-of-distribution (OOD) effects?}
The distribution mismatch refers to the situation when train and test sets come from different distributions. Usually, training data used to develop the model does not entirely represent the real-world data (test set) it will eventually encounter. An example would be urban training areas for autonomous driving while the vehicle is going to be tested in off-the-road regions with challenges outside of the distribution of urban areas. In the context of interpretability, however, the methods are expected to accurately identify important features within the input data that cover the challenges to contribute to the network's classification. while there might be an apparent mismatch between the perturbed training set and the unperturbed test set, it can be deduced that ODD will not occur as long as the network shows generalization. Thus, in this study, we only focus on analyzing attribution methods when they show generalizability. By doing so, we circumvent the above-mentioned problems related to a perturbed test set and also have taken into account the OOD effect.

\begin{figure}[!ht]
    \centering

    \subfloat[Mutag]{%
      \includegraphics[clip,width=0.7\columnwidth]{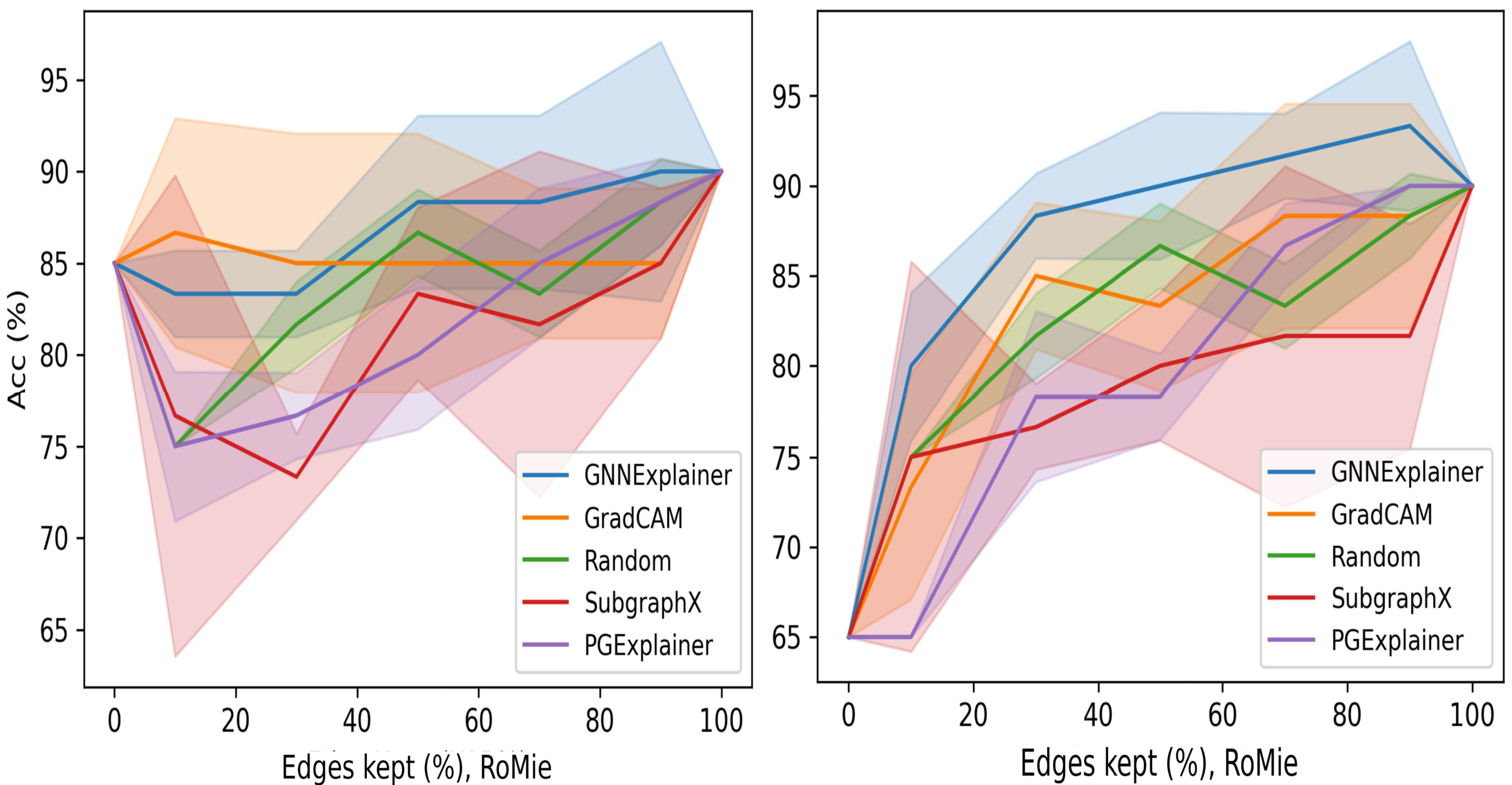}%
    }%

    \subfloat[Enzyme]{%
      \includegraphics[clip,width=0.7\columnwidth]{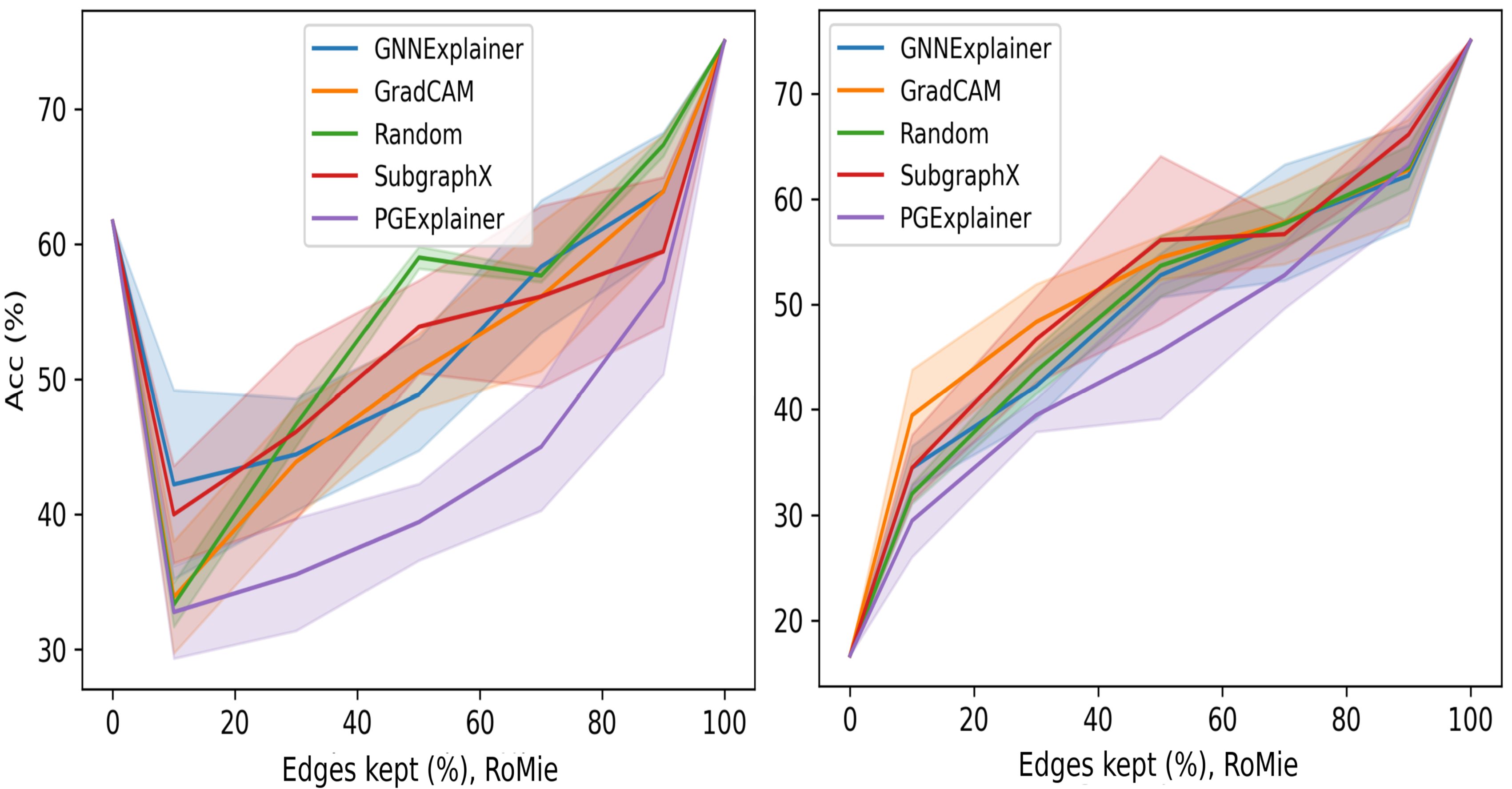}%
    }

    \caption{\textbf{Elimination of Isolated Nodes:}
    {\textbf{with}} (left), and {\textbf{without}} (right) isolated nodes.
    Isolated atoms do not preserve any properties for a molecule, isolated nodes artificially cause high accuracy at low sparsity.
    {\textbf{a}} on Mutag, and {\textbf{b}} on Enzyme, both for GCN
    }
   \label{fig4:nodeEliReal}
   
\end{figure}

\subsection{Elimination of Isolated Nodes}
\noindent After edge removal by the attribution method, certain nodes may persist in the subgraph in isolation, i.e., without any connectivity to other nodes. While these isolated nodes lack neighboring nodes and thus will not have an influence on their aggregation and updating steps, they may still impact the network. In graph datasets where nodes possess features, the impact of individual nodes on the network's performance regardless of the neighbors is expected. As a result, even though the isolated nodes do not play a role in updating other nodes, their inherent features still affect the overall network dynamics.

\begin{figure}[!ht]
    \centering

    \subfloat[Keeping Isolated Nodes]{%
      \includegraphics[clip,width=0.7\columnwidth]{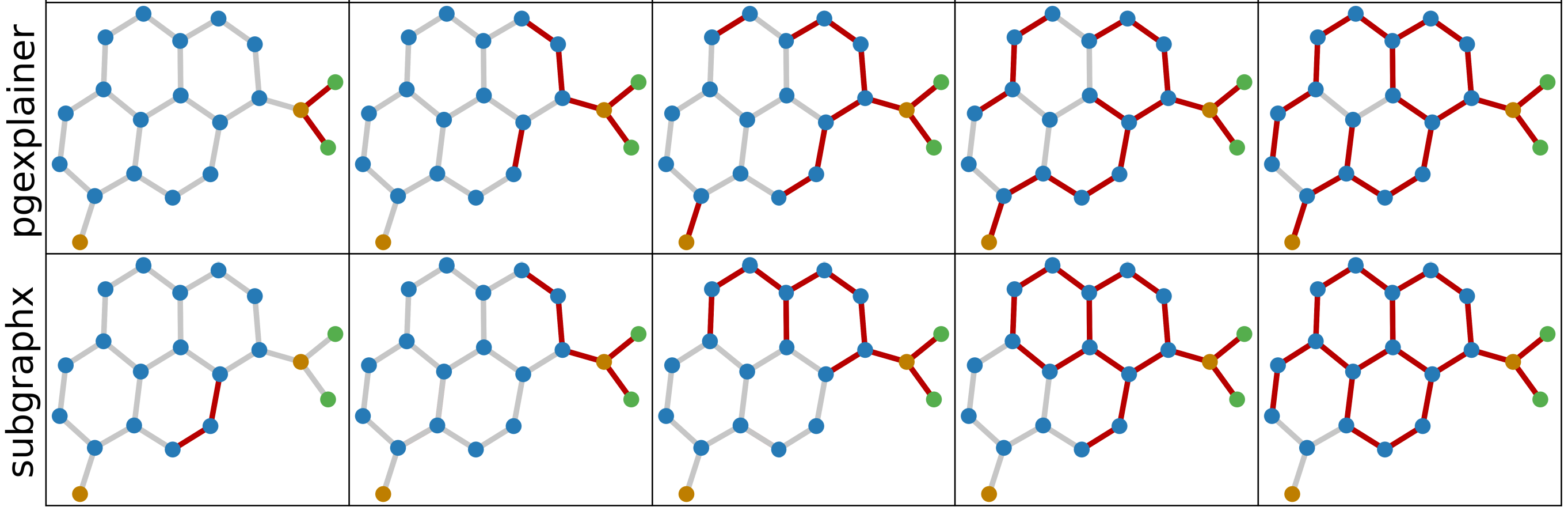}%
    }%

    \subfloat[Elimination Isolated Nodes]{%
      \includegraphics[clip,width=0.7\columnwidth]{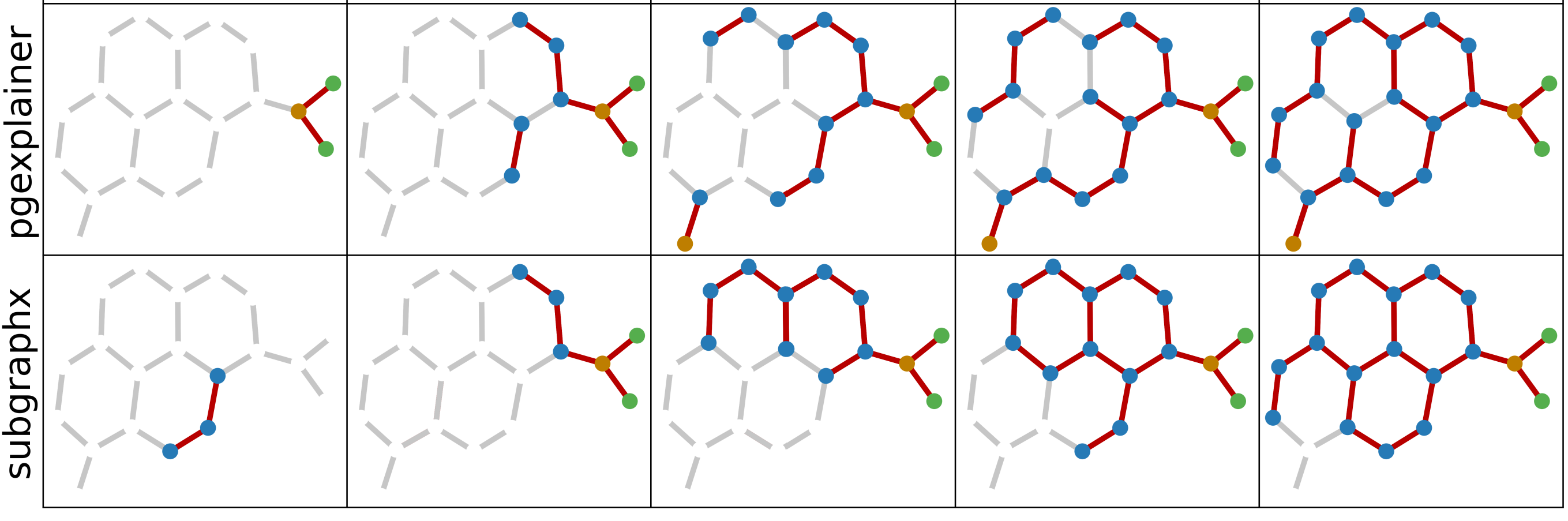}%
    }
    
    \caption{\textbf{Sample Visialization for Isolated Node Elimination:}.
    When isolated nodes are not eliminated, although there is no edge at low sparsity, still high accuracy is obtained due to lasting node features. on Mutag, for GCN.
    }
   \label{fig5:nodeEliVisMutag}
   
\end{figure}

The impact of node elimination when nodes do not possess features, and therefore have no predictive information, can be observed in Figure~\ref{fig3:nodeEliSynth}. We show the difference between RoMie applied on REDDIT-BINARY and BA3Motifs datasets with the absence or presence of isolated nodes. No significant disparity was observed. Hence, in datasets with non-existent node features, the elimination of isolated nodes has negligible influence on the retraining process. It is not the case for datasets with node features, however. 
As demonstrated in Figure~\ref{fig4:nodeEliReal} and Figure~\ref{fig5:nodeEliVisMutag}, 
when applying RoMie to MUTAG and ENZYME, the isolated nodes exhibit discriminative capabilities. Including these isolated nodes during retraining leads to a relatively high network accuracy at low sparsity, $65\%$, and $85\%$, respectively. This suggests that despite keeping only a few of the most important edges, the presence of isolated nodes reinforces the network. Conversely, when these isolated nodes are eliminated, the accuracy drops to $17\%$ for MUTAG and $65\%$ for ENZYME, showing artificially high network accuracy.

Based on this observation, we propose to eliminate any isolated nodes that arise after the attribution-finding process: Firstly, in the context of chemical and biological real-world datasets, a subgraph is meaningful upon the interconnectivity of all its consisting nodes. An isolated, non-bonding atom in a molecule or enzyme is not sensible. Secondly, the primary objective of our analysis centers on discovering the importance of edges. A node turns isolated because all its edges are already moved, therefore the edge removal process implies the removal of related isolated nodes emerging during this process as well.

\subsection{Implementation Details}
All attributions provide probability edge weighting. Therefore, the attribution is the subset of the graph selected based on the edge weights. However, SubgraphX does not provide probability weights for edges. Therefore we generated the binary edge mask for each percentage in our KAR and ROAR experiments. E.g., for the x\%, we search for a x\% size subgraph. We perform retraining by applying edge removing/keeping using probability importance weightings provided by the attributions, albeit as mentioned before, for SubgraphX, we directly compute the subgraph. In general, we have implemented our experiments based on seven percentages of 0, 10, 30, 50, 70, 90, and 100 defined by the number of removing/keeping edges. During random experiments, each one is performed three times with a separate set of random probability weights with different seeds.

\section{Discussions}

\begin{figure}[!ht]
    \centering

    \subfloat[RoMie and RoLie Bahavior]{%
      \includegraphics[clip,width=0.7\columnwidth]{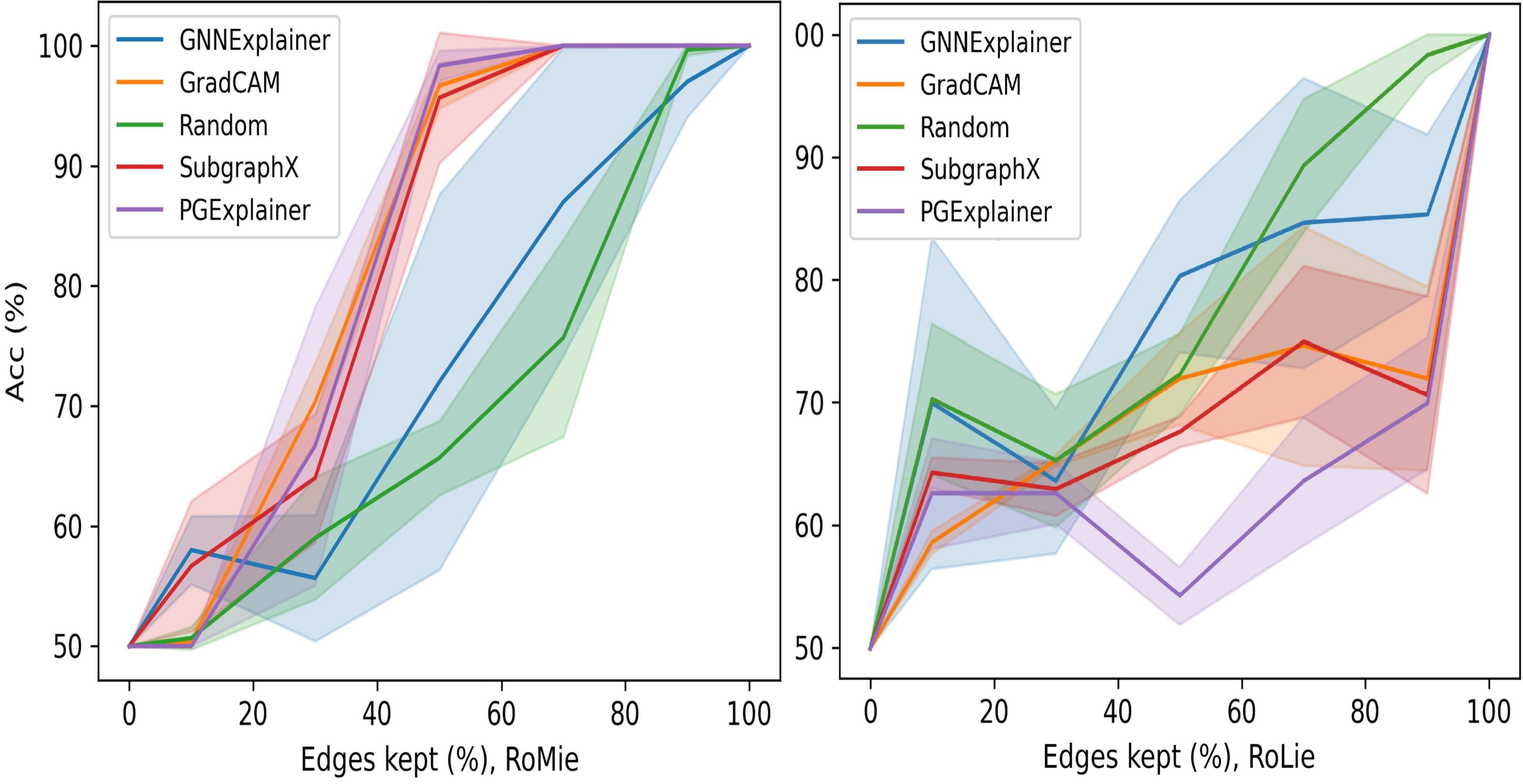}%
    }%

    \subfloat[Label zero (cycle)]{%
      \includegraphics[clip,width=0.7\columnwidth]{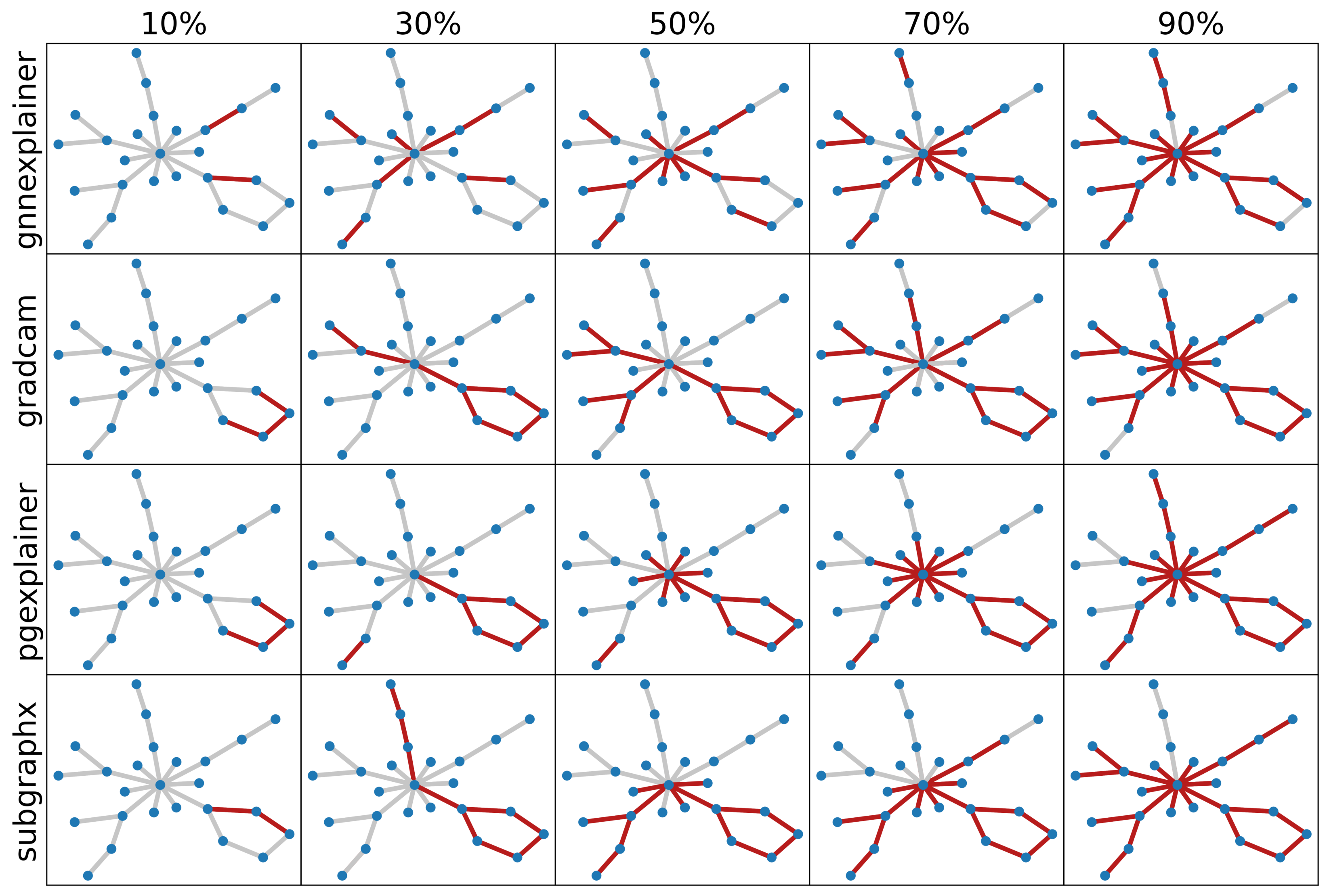}%
    }%

    \subfloat[Label one (house)]{%
      \includegraphics[clip,width=0.7\columnwidth]{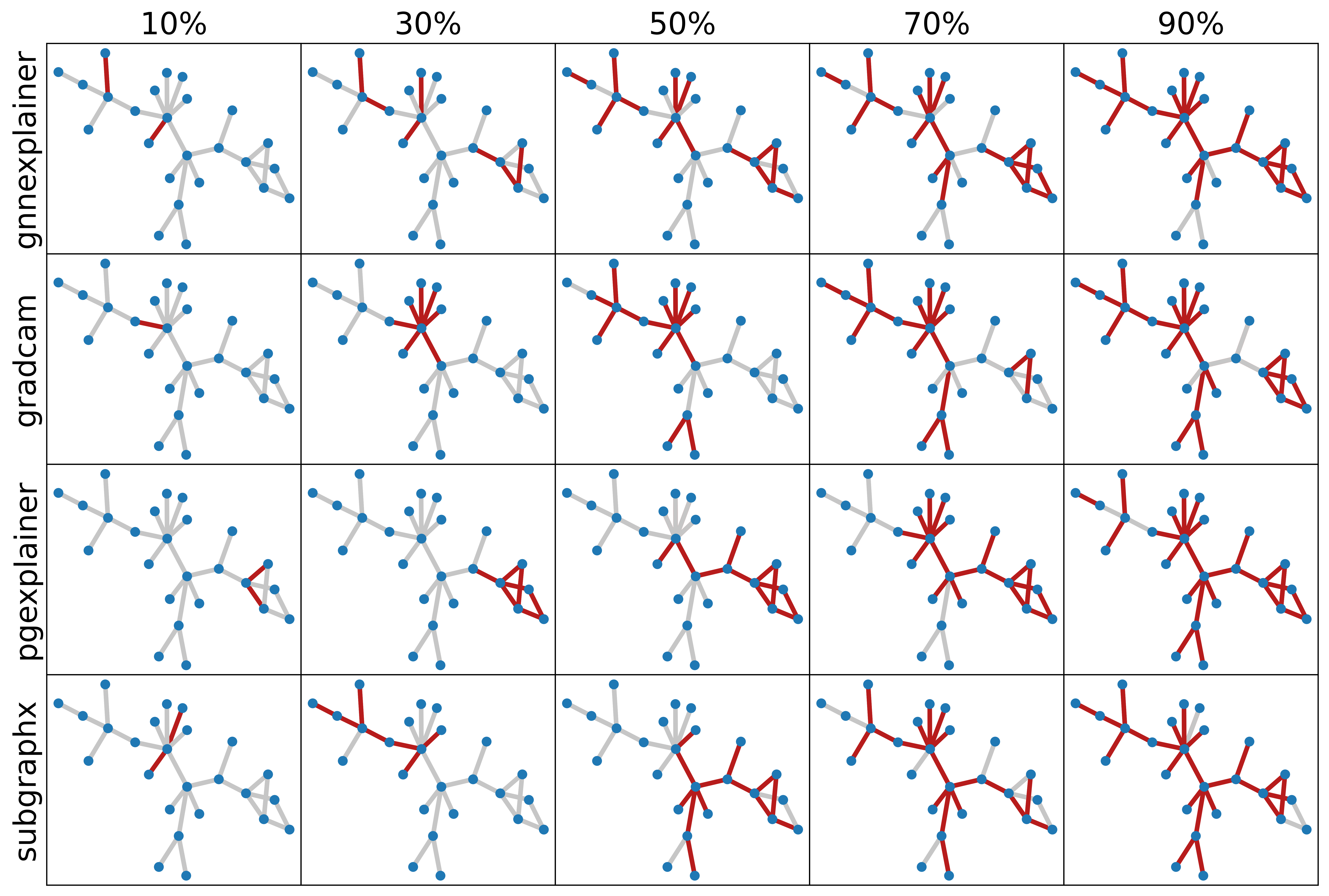}%
    }

    \caption{\textbf{Complementary RoMie and RoLie:}
    Important and Least important edges should be studied together. A visualization of attribution can help understand the inner workings of a method. on BA2Motif, for GCN.}
   \label{fig6:RoMieRoLie}
   
\end{figure}

\subsection{RoMie and RoLie are Complementary}
Earlier we mentioned it is insufficient to base the attribution evaluation solely on one retraining strategy and parallel evaluation of Most and Least important edges is crucial. \textit{Four} scenarios come into play when looking at methods' performance across different sparsities:

\textit{ \textbf{i)}} When there is a sharp rise in low levels of sparsity for RoMie, the behavior of RoLie is a pivotal factor:
\textit{ \textbf{a) if RoLie also displays a sharp rise at the beginning}}, the overall method seems to behave unpredictably, lacking a discernible pattern. We could see this in GradCAM in Figure~\ref{fig6:RoMieRoLie}. Accuracy in the RoMie curve quickly rises to the original accuracy. However, as the method ignores some discriminative edges, during RoLie experiments, some discriminative edges keep the accuracy high, and thus GradCAM is undesirable from this perspective. From the visual explanations in Figure~\ref{fig6:RoMieRoLie} (b) and (c), we can relate it to the fact that GradCam ignores either cycle or house structure in RoMie, and therefore will be present in the least important edges with discrimination power for the network. On the other hand, \textit{ \textbf{b) if RoLie exhibits a smooth rise}}, it suggests the method's ability for generalization. PGExplainer and SubgraphX both show this behavior in the two retraining settings. With a look at the visual examples, we can see that PGExplainer captures the whole part of both cycle and house structures, giving us a piece of strong evidence that GCN potentially focuses on these features during its retraining. SubgraphX returns the most but not the entire part of the house pattern, which has a high overlapping part with the cycle one. Hence, removing them could disrupt network training. Therefore, it does not show a bad performance in the ROAR experiment.

\textit{ \textbf{ii)}} In cases where the early sparsity ascent for RoMie lacks sharpness, likewise,  RoLie's characteristics further determine the method's behavior: \textit{ \textbf{a) if RoLie remains sharp in such a context}}, it signals an inability to generalize. In Figure~\ref{fig6:RoMieRoLie} we see that GNNExplainer nearly performs on par with the random setting in both retraining evaluations, and its visualized outputs demonstrate a lack of confidence behind both cycle and house patterns. However, \textit{ \textbf{b) in case of a smooth RoLie}} accompanied by a method that appears random implies a lack of consistency. The method does not pick the most important edges at low sparsities, yet it does not leave them behind in the least important list. This is a case where the model is not very smart in identifying attributions. For BA2Motif dataset, non of the methods exhibited this behavior. For such examples please look at supplementary.

Thus, the takeaway here is that the merit of an attribution method is not merely determined by the success of one retraining strategy; RoMie and RoLie must be analyzed jointly for a well-rounded and sound judgment.

\subsection{Explainers Depend on Datasets and Networks}
A central point of concern in the evaluation of attributions could revolve around whether the behavior of methods remains consistent when applied to different datasets and network architectures, or if their effectiveness in terms of generalizability varies from one setting to another. In Figure~\ref{fig6:RoMieRoLie}, we employed the GCN and evaluated the outcomes of retraining. Shifting our focus to Figure~\ref{fig8:BA3}, we aim to trace the results of the four methods on the same dataset (BA2Motif), but this time employing the GIN architecture. This as a comparison point allows us to monitor the influence of a different network on the outcomes.

Previously for GCN, we observed PGExplainer can precisely identify circle-house motif pair in low sparsity. This implied the edges related to the motif would not appear among the least important edges. That is why a smooth RoLie was observed. For the GIN
(shown in Supplementary),
PGExplainer overlooks either Cycle or house motif. This will let some edges related to the motifs remain in the subgraph when retraining on the least importants. Therefore, it will lead to undesirable high accuracy at low sparsity for RoLie.

\begin{figure}[!ht]
    \centering

    \subfloat[GIN: RoMie and RoLie]{%
      \includegraphics[clip,width=0.7\columnwidth]{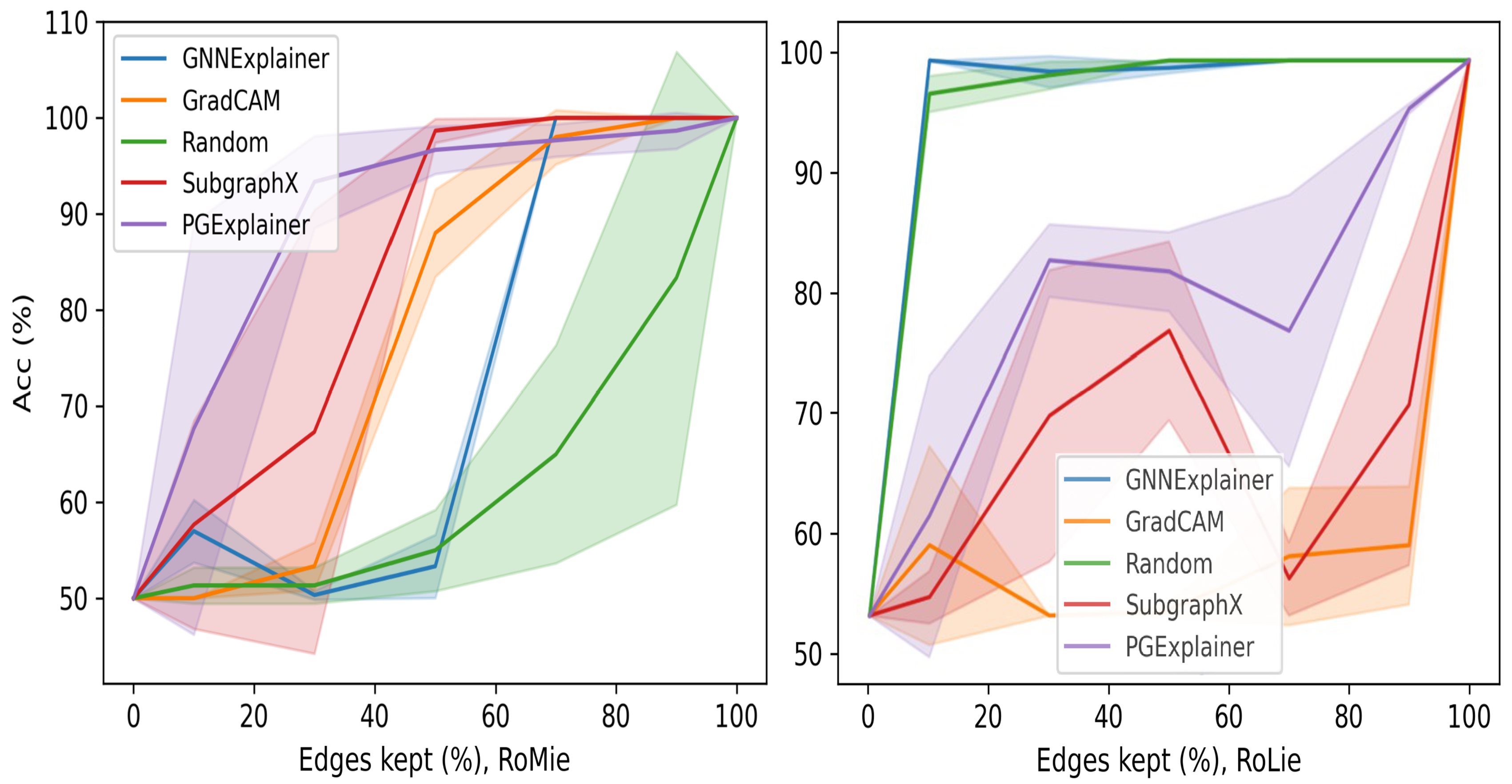}%
    }%

    \subfloat[BA3Motif: RoMie and RoLie]{%
      \includegraphics[clip,width=0.7\columnwidth]{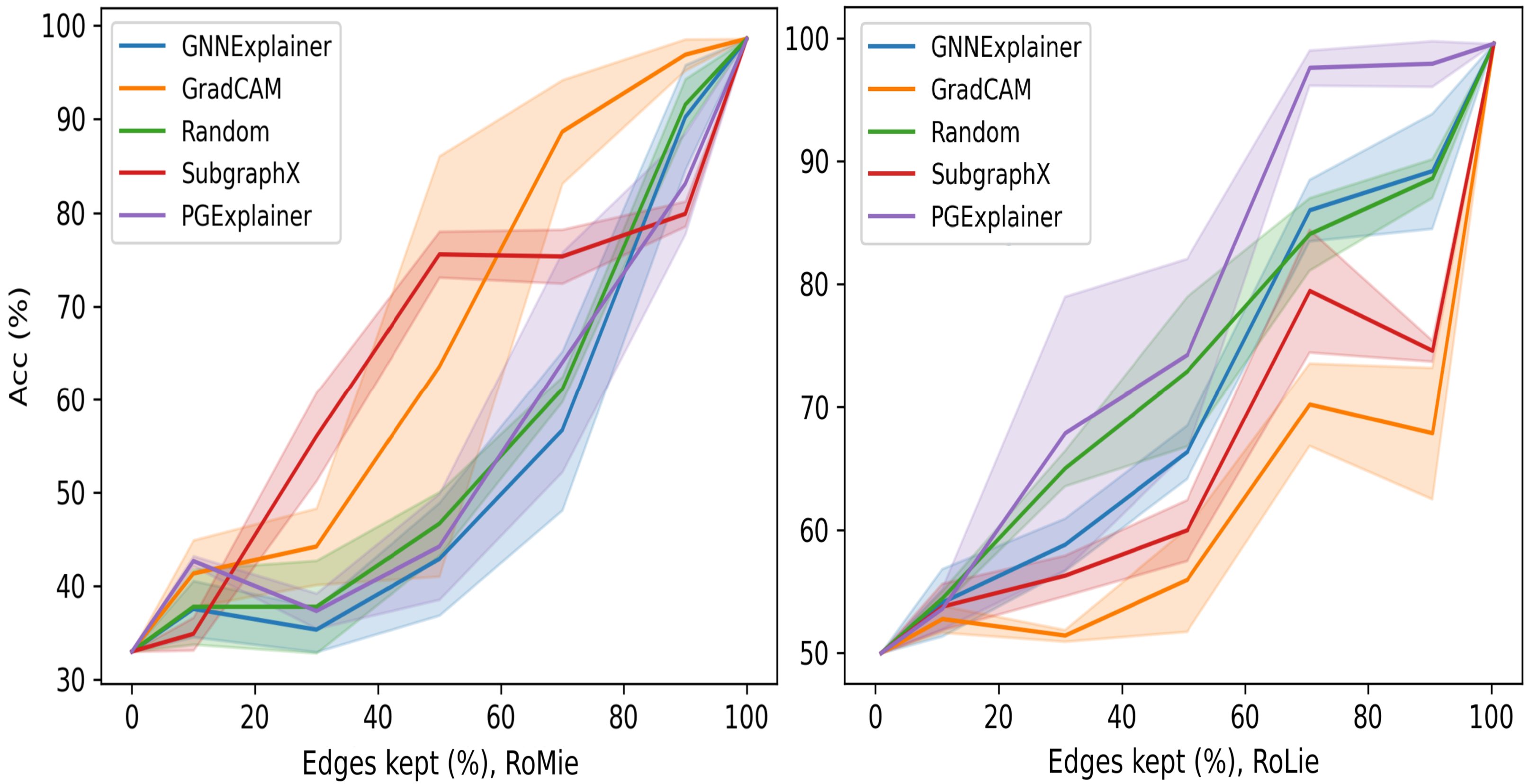}%
    }%

    \caption{\textbf{Attribitions Dependency on Network} (up) \textbf{and Dataset} (down). When compared to \ref{fig6:RoMieRoLie} the performance of attributions changes. Up: on BA2Motif, for GIN. Down: on BA3Motif, for GCN}
   \label{fig7:BA3}
   
\end{figure}

\begin{figure}[!ht]
    \centering

    \subfloat[Label zero ()]{%
      \includegraphics[clip,width=0.7\columnwidth]{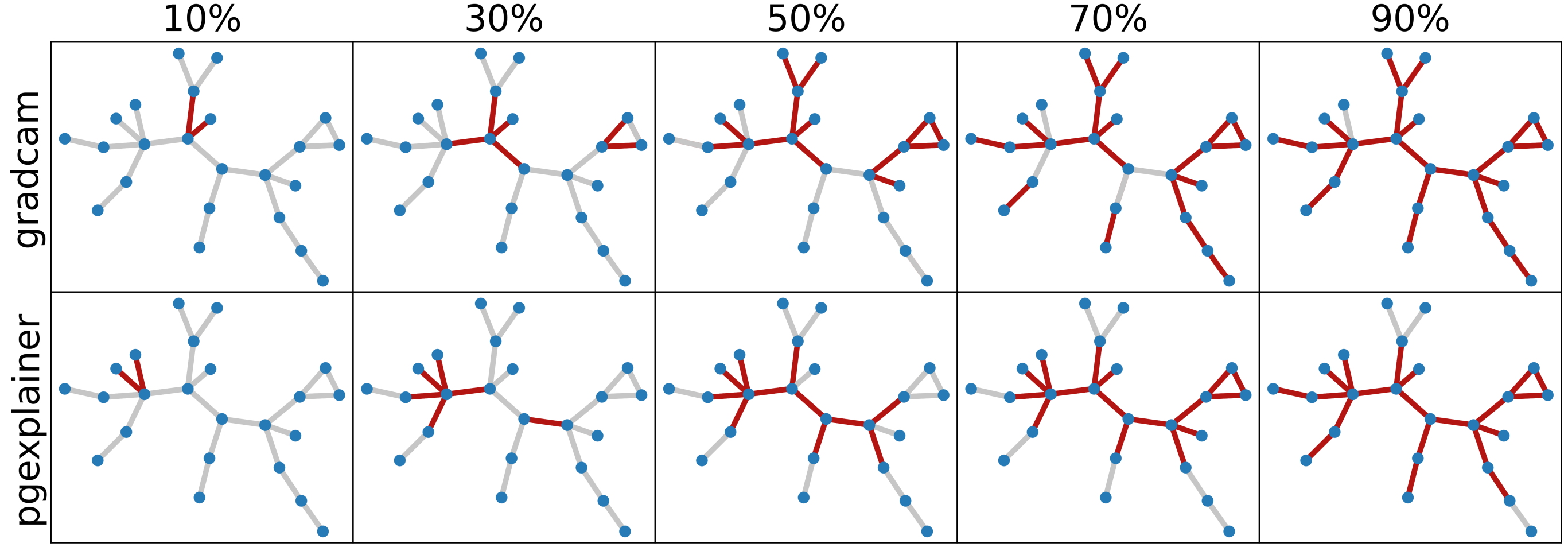}%
    }

    \caption{\textbf{Visualization of a Sample with Triangle}.
    When the dataset is different from \ref{fig6:RoMieRoLie} the performance of attribution methods also changes. on BA3Motif, for GCN}
   \label{fig8:BA3}
   
\end{figure}

Unlike PGExplainer, GradCAM captures the whole Cycle-house pair in the new GIN setting. In Figure~\ref{fig6:RoMieRoLie} (b) and (c) we noted GradCAM was unable to capture house motifs in contrast to Figure~\ref{fig8:BA3} (b) and (c) where we see the pair is discernable for the method. Overall, results for GradCam look fairly consistent in both architectures. This is also the case for GNNExplainer and SubgraphX. GNNExplainer continues to exhibit similar behavior close to random as before. We only observe an increase in accuracy for it once the sparsity level reaches 50\%. What sets this method different in the context of GIN is that it shows a wide range of variability, making it less robust in the GIN setting. For subgraphX in RoLie plot, we observe a peculiar behavior. There is a sudden jump in the curve. This behavior can be justified by the fact that SubgraphX results do not necessarily overlap if we increase the search space (e.g., from 30\% to 50\%). All in all, results in Figure~\ref{fig7:BA3} prove the inconsistency of methods when employed across different networks.

\textbf{From Binary to Trinary Classification:} 
Binary datasets like BA2Motifs have a notable limitation. Achieving high accuracy is feasible by focusing on a single discriminative motif and disregarding the features of the other motif. To address this, we introduce BA3Motifs, an extended multi-class variant of BA2Motifs, where the triangle motif replaces the house motif to create a third class. Figure~\ref{fig8:BA3} shows experiments to evaluate attribution method's data dependency using this dataset. In contrast to BA2Motifs, retaining PGExplainer features has no positive impact on network retraining, showing a performance similar to random settings. Retraining the least important features also does not reduce the performance. This is because PGExplainer fails to capture either cycle, house, or triangle motifs, as can be seen in the sample visualization. This evidence indicates that methods such as PGExplainer might exhibit inconsistent behavior in BA2Motifs and BA3Motifs, implying that the method's performance is contingent on the dataset.

\subsection{Which Method to Recommend?}
So far, our observations have highlighted the dependence of attribution methods on both the dataset and the network. The next question that arises is whether, in a specific setting, we can make a general recommendation for a method. Our attempts in this study were not directed at creating a benchmark; rather, our aim was to introduce a framework for fair evaluation. However, as we employed the RoMie and RoLie approaches across various settings, evident trends could be seen:
PGExplainer shows to be a method that has the most variable performance among different settings, while GradCAM is relatively more stable. GNNExplainer was the method that performed similarly to the random setting in most experiences. SubgraphX shows the disadvantage of non-overlapping subgraphs as sparsity was increasing. For instance, in sample visualization as in Figure~\ref{fig1:teaser} for the BA2Motif, the selected nodes in sparsity level of $30\%$ were not necessarily appearing also in $70\%$. This can be associated with the fact that subgraphX has a different search span each time it looked for attributes. This lesd to huge jumps both in RoMie and RoLie plots. Therefore we recommend using subgraphX for datasets of small graph size where the search space can be limited. 
In addition to the trend, a key observation can be drawn from both the RoMie and RoLie plots: at a specific sparsity, a method can demonstrate superior performance compared to another one, while underperforming the same method at a different sparsity level. Keeping this in mind, it becomes essential to initially determine the desired sparsity level and then proceed to method comparison. What holds particularly important here is the alignment of the sparsity level of $x\%$ in RoMie with the sparsity level of $(100-x)\%$ in RoLie, as these two subsets Combined represent the complete original graph sample $G_o$. Thus, for one method to be considered more reliable than another, it must exhibit higher accuracy in RoMie and lower accuracy in the complementary sparsity within RoLie. When coupled with the visualization of attributions, these insights provide us with a robust basis for comparison.

\section{Conclusion}
In this work, we introduce guidelines on how to objectively evaluate graph attributions through the perspective of retraining. We reformulate the previous retraining paradigm with a focus on issues regarding the train-test distribution mismatch. The evaluations reveal that a mainstream method, GNNExplainer, is consistently failing to highlight predictive attributions. The idea of benchmarking attribution methods is not reliable, as they usually show different behaviors on different datasets, networks, and also at different levels of sparsity. Therefore we recommend first specifying a clear problem condition and then using the proposed evaluation strategies for a fair comparison of attribution methods.

\bibliography{aaai24}

\begin{thebibliography}{34}
\providecommand{\natexlab}[1]{#1}

\bibitem[{Adebayo et~al.(2018)Adebayo, Gilmer, Muelly, Goodfellow, Hardt, and
  Kim}]{adebayo2018sanity}
Adebayo, J.; Gilmer, J.; Muelly, M.; Goodfellow, I.; Hardt, M.; and Kim, B.
  2018.
\newblock Sanity checks for saliency maps.
\newblock \emph{Advances in neural information processing systems}, 31.

\bibitem[{Agarwal, Zitnik, and Lakkaraju(2022)}]{agarwal2022probing}
Agarwal, C.; Zitnik, M.; and Lakkaraju, H. 2022.
\newblock Probing gnn explainers: A rigorous theoretical and empirical analysis
  of gnn explanation methods.
\newblock In \emph{International Conference on Artificial Intelligence and
  Statistics}, 8969--8996. PMLR.

\bibitem[{Bach et~al.(2015)Bach, Binder, Montavon, Klauschen, M{\"u}ller, and
  Samek}]{cv2}
Bach, S.; Binder, A.; Montavon, G.; Klauschen, F.; M{\"u}ller, K.-R.; and
  Samek, W. 2015.
\newblock On pixel-wise explanations for non-linear classifier decisions by
  layer-wise relevance propagation.
\newblock \emph{PloS one}, 10(7): e0130140.

\bibitem[{Dabkowski and Gal(2017)}]{cv1}
Dabkowski, P.; and Gal, Y. 2017.
\newblock Real time image saliency for black box classifiers.
\newblock \emph{Advances in neural information processing systems}, 30.

\bibitem[{Debnath et~al.(1991)Debnath, Lopez~de Compadre, Debnath, Shusterman,
  and Hansch}]{debnath1991structure}
Debnath, A.~K.; Lopez~de Compadre, R.~L.; Debnath, G.; Shusterman, A.~J.; and
  Hansch, C. 1991.
\newblock Structure-activity relationship of mutagenic aromatic and
  heteroaromatic nitro compounds. correlation with molecular orbital energies
  and hydrophobicity.
\newblock \emph{Journal of medicinal chemistry}, 34(2): 786--797.

\bibitem[{Fong and Vedaldi(2017)}]{cv3}
Fong, R.~C.; and Vedaldi, A. 2017.
\newblock Interpretable explanations of black boxes by meaningful perturbation.
\newblock In \emph{Proceedings of the IEEE international conference on computer
  vision}, 3429--3437.

\bibitem[{Hooker et~al.(2019)Hooker, Erhan, Kindermans, and Kim}]{ROAR}
Hooker, S.; Erhan, D.; Kindermans, P.-J.; and Kim, B. 2019.
\newblock A benchmark for interpretability methods in deep neural networks.
\newblock \emph{Advances in neural information processing systems}, 32.

\bibitem[{Jang, Gu, and Poole(2016)}]{pgExp_reparTrick}
Jang, E.; Gu, S.; and Poole, B. 2016.
\newblock Categorical reparameterization with gumbel-softmax.
\newblock \emph{arXiv preprint arXiv:1611.01144}.

\bibitem[{Khakzar et~al.(2021)Khakzar, Baselizadeh, Khanduja, Rupprecht, Kim,
  and Navab}]{Khakzar_2021_CVPR}
Khakzar, A.; Baselizadeh, S.; Khanduja, S.; Rupprecht, C.; Kim, S.~T.; and
  Navab, N. 2021.
\newblock Neural Response Interpretation Through the Lens of Critical Pathways.
\newblock In \emph{Proceedings of the IEEE/CVF Conference on Computer Vision
  and Pattern Recognition (CVPR)}, 13528--13538.

\bibitem[{Khakzar et~al.(2022)Khakzar, Khorsandi, Nobahari, and
  Navab}]{khakzar2022explanations}
Khakzar, A.; Khorsandi, P.; Nobahari, R.; and Navab, N. 2022.
\newblock Do explanations explain? model knows best.
\newblock In \emph{Proceedings of the IEEE/CVF Conference on Computer Vision
  and Pattern Recognition}, 10244--10253.

\bibitem[{Kipf and Welling(2016)}]{GCN}
Kipf, T.~N.; and Welling, M. 2016.
\newblock Semi-supervised classification with graph convolutional networks.
\newblock \emph{arXiv preprint arXiv:1609.02907}.

\bibitem[{Krishna et~al.(2022)Krishna, Han, Gu, Pombra, Jabbari, Wu, and
  Lakkaraju}]{krishna2022disagreement}
Krishna, S.; Han, T.; Gu, A.; Pombra, J.; Jabbari, S.; Wu, S.; and Lakkaraju,
  H. 2022.
\newblock The Disagreement Problem in Explainable Machine Learning: A
  Practitioner's Perspective.
\newblock \emph{arXiv preprint arXiv:2202.01602}.

\bibitem[{Kuhn and Tucker(1953)}]{subGraphX_shapley_1}
Kuhn, H.~W.; and Tucker, A.~W. 1953.
\newblock \emph{Contributions to the Theory of Games}.
\newblock 28. Princeton University Press.

\bibitem[{Li et~al.(2022)Li, Yang, Pagnucco, and Song}]{li2022explainability}
Li, P.; Yang, Y.; Pagnucco, M.; and Song, Y. 2022.
\newblock Explainability in Graph Neural Networks: An Experimental Survey.
\newblock \emph{arXiv preprint arXiv:2203.09258}.

\bibitem[{Liu et~al.(2021)Liu, Luo, Wang, Xie, Yuan, Gui, Yu, Xu, Zhang, Liu
  et~al.}]{liu2021dig}
Liu, M.; Luo, Y.; Wang, L.; Xie, Y.; Yuan, H.; Gui, S.; Yu, H.; Xu, Z.; Zhang,
  J.; Liu, Y.; et~al. 2021.
\newblock Dig: A turnkey library for diving into graph deep learning research.
\newblock \emph{arXiv preprint arXiv:2103.12608}.

\bibitem[{Lundberg and Lee(2017)}]{subGraphX_shapley_2}
Lundberg, S.~M.; and Lee, S.-I. 2017.
\newblock A unified approach to interpreting model predictions.
\newblock \emph{Advances in neural information processing systems}, 30.

\bibitem[{Luo et~al.(2020)Luo, Cheng, Xu, Yu, Zong, Chen, and Zhang}]{pgExp}
Luo, D.; Cheng, W.; Xu, D.; Yu, W.; Zong, B.; Chen, H.; and Zhang, X. 2020.
\newblock Parameterized explainer for graph neural network.
\newblock \emph{Advances in neural information processing systems}, 33:
  19620--19631.

\bibitem[{Maddison, Mnih, and Teh(2016)}]{pgExp_binary}
Maddison, C.~J.; Mnih, A.; and Teh, Y.~W. 2016.
\newblock The concrete distribution: A continuous relaxation of discrete random
  variables.
\newblock \emph{arXiv preprint arXiv:1611.00712}.

\bibitem[{Montavon et~al.(2017)Montavon, Lapuschkin, Binder, Samek, and
  M{\"u}ller}]{cv4}
Montavon, G.; Lapuschkin, S.; Binder, A.; Samek, W.; and M{\"u}ller, K.-R.
  2017.
\newblock Explaining nonlinear classification decisions with deep taylor
  decomposition.
\newblock \emph{Pattern recognition}, 65: 211--222.

\bibitem[{Pope et~al.(2019)Pope, Kolouri, Rostami, Martin, and
  Hoffmann}]{gradCam}
Pope, P.~E.; Kolouri, S.; Rostami, M.; Martin, C.~E.; and Hoffmann, H. 2019.
\newblock Explainability methods for graph convolutional neural networks.
\newblock In \emph{Proceedings of the IEEE/CVF Conference on Computer Vision
  and Pattern Recognition}, 10772--10781.

\bibitem[{Ribeiro, Singh, and Guestrin(2016)}]{trust}
Ribeiro, M.~T.; Singh, S.; and Guestrin, C. 2016.
\newblock "Why should i trust you?" Explaining the predictions of any
  classifier.
\newblock In \emph{Proceedings of the 22nd ACM SIGKDD international conference
  on knowledge discovery and data mining}, 1135--1144.

\bibitem[{Samek et~al.(2016)Samek, Binder, Montavon, Lapuschkin, and
  M{\"u}ller}]{samek2016evaluating}
Samek, W.; Binder, A.; Montavon, G.; Lapuschkin, S.; and M{\"u}ller, K.-R.
  2016.
\newblock Evaluating the visualization of what a deep neural network has
  learned.
\newblock \emph{IEEE transactions on neural networks and learning systems},
  28(11): 2660--2673.

\bibitem[{Sanchez-Lengeling et~al.(2020)Sanchez-Lengeling, Wei, Lee, Reif,
  Wang, Qian, McCloskey, Colwell, and Wiltschko}]{sanchez2020evaluating}
Sanchez-Lengeling, B.; Wei, J.; Lee, B.; Reif, E.; Wang, P.; Qian, W.;
  McCloskey, K.; Colwell, L.; and Wiltschko, A. 2020.
\newblock Evaluating attribution for graph neural networks.
\newblock \emph{Advances in neural information processing systems}, 33:
  5898--5910.

\bibitem[{Schomburg et~al.(2004)Schomburg, Chang, Ebeling, Gremse, Heldt, Huhn,
  and Schomburg}]{schomburg2004brenda}
Schomburg, I.; Chang, A.; Ebeling, C.; Gremse, M.; Heldt, C.; Huhn, G.; and
  Schomburg, D. 2004.
\newblock BRENDA, the enzyme database: updates and major new developments.
\newblock \emph{Nucleic acids research}, 32(suppl\_1): D431--D433.

\bibitem[{Selvaraju et~al.(2017)Selvaraju, Cogswell, Das, Vedantam, Parikh, and
  Batra}]{gradCam_Vision}
Selvaraju, R.~R.; Cogswell, M.; Das, A.; Vedantam, R.; Parikh, D.; and Batra,
  D. 2017.
\newblock Grad-cam: Visual explanations from deep networks via gradient-based
  localization.
\newblock In \emph{Proceedings of the IEEE international conference on computer
  vision}, 618--626.

\bibitem[{V{\"o}lske et~al.(2017)V{\"o}lske, Potthast, Syed, and
  Stein}]{reddit}
V{\"o}lske, M.; Potthast, M.; Syed, S.; and Stein, B. 2017.
\newblock {TL};{DR}: Mining {R}eddit to Learn Automatic Summarization.
\newblock In \emph{Proceedings of the Workshop on New Frontiers in
  Summarization}, 59--63. Copenhagen, Denmark: Association for Computational
  Linguistics.

\bibitem[{Xu et~al.(2018)Xu, Hu, Leskovec, and Jegelka}]{GIN}
Xu, K.; Hu, W.; Leskovec, J.; and Jegelka, S. 2018.
\newblock How powerful are graph neural networks?
\newblock \emph{arXiv preprint arXiv:1810.00826}.

\bibitem[{Ying et~al.(2019)Ying, Bourgeois, You, Zitnik, and Leskovec}]{gnnExp}
Ying, Z.; Bourgeois, D.; You, J.; Zitnik, M.; and Leskovec, J. 2019.
\newblock Gnnexplainer: Generating explanations for graph neural networks.
\newblock \emph{Advances in neural information processing systems}, 32.

\bibitem[{Yuan et~al.(2020)Yuan, Tang, Hu, and Ji}]{yuan2020xgnn}
Yuan, H.; Tang, J.; Hu, X.; and Ji, S. 2020.
\newblock Xgnn: Towards model-level explanations of graph neural networks.
\newblock In \emph{Proceedings of the 26th ACM SIGKDD International Conference
  on Knowledge Discovery \& Data Mining}, 430--438.

\bibitem[{Yuan et~al.(2022)Yuan, Yu, Gui, and Ji}]{gnn_interpretability_Survey}
Yuan, H.; Yu, H.; Gui, S.; and Ji, S. 2022.
\newblock Explainability in graph neural networks: A taxonomic survey.
\newblock \emph{IEEE Transactions on Pattern Analysis and Machine
  Intelligence}.

\bibitem[{Yuan et~al.(2021)Yuan, Yu, Wang, Li, and Ji}]{subGraphX}
Yuan, H.; Yu, H.; Wang, J.; Li, K.; and Ji, S. 2021.
\newblock On explainability of graph neural networks via subgraph explorations.
\newblock In \emph{International Conference on Machine Learning}, 12241--12252.
  PMLR.

\bibitem[{Zhang et~al.(2018)Zhang, Bargal, Lin, Brandt, Shen, and
  Sclaroff}]{zhang2018top}
Zhang, J.; Bargal, S.~A.; Lin, Z.; Brandt, J.; Shen, X.; and Sclaroff, S. 2018.
\newblock Top-down neural attention by excitation backprop.
\newblock \emph{International Journal of Computer Vision}, 126(10): 1084--1102.

\bibitem[{Zhang et~al.(2021)Zhang, Khakzar, Li, Farshad, Kim, and
  Navab}]{zhang2021fine}
Zhang, Y.; Khakzar, A.; Li, Y.; Farshad, A.; Kim, S.~T.; and Navab, N. 2021.
\newblock Fine-grained neural network explanation by identifying input features
  with predictive information.
\newblock \emph{Advances in Neural Information Processing Systems}, 34:
  20040--20051.

\bibitem[{Zhang et~al.(2023)Zhang, Li, Brown, Rezaei, Bischl, Torr, Khakzar,
  and Kawaguchi}]{zhang2023attributionlab}
Zhang, Y.; Li, Y.; Brown, H.; Rezaei, M.; Bischl, B.; Torr, P.; Khakzar, A.;
  and Kawaguchi, K. 2023.
\newblock AttributionLab: Faithfulness of Feature Attribution Under
  Controllable Environments.
\newblock arXiv:2310.06514.

\end{thebibliography}

\clearpage
\textbf{Supplementary Material:}
\section{Experiments' Details}

\subsection{Networks and Training}
Initially, for each dataset a three-layer GIN and GCN were trained; all three layers have an identical number of neurons though their number might vary from 20 to 300 depending on the dataset and network. BA-2Motifs was trained for 100 epochs with batch size 64 for GCN and 32 for GIN, BA-3Motifs for 100 epochs with batch size 64, REDDIT-BINARY, IMDB-BINARY, ENZYME, and MTAG for 500 epochs and early stopping of 100 with batch size 64, though for MUTAG batch size was 32. Adam optimizer with a learning rate of 0.001 was employed. We applied the same configuration during retraining strategies i.e. RoLie and RoLie.
\

\subsection{Hardware}
We used two Nvidia GPUs ”Quadro P6000” and ”GeForce GTX 1080 Ti” on an ”AMD Ryzen 7 1080X Eight-Core Processor” CPU, Python version 3.10.4, Pytorch 1.12, Torch-geometric 1.7.2 and Dive-Into-Graphs (DIG) 0.1.2 packages.

\subsection{Datasets}
\begin{center}
\begin{tabular}{ |c|c|c|c| } 
\hline
Dataset & \# graphs & avg. \# of nodes & \# edges\\
\hline

MUTAG      & 188   &   17.93  &  19.79  \\
ENZYME     & 600   &   32.63  &   62.14 \\
IMDB-BIN   & 1000  &   19.77  &  96.53  \\
REDDIT-BIN & 2000  &   429.63 &  497.75 \\
BA-2Motifs & 1000  &   25     &  25.48  \\
BA-3Motifs & 1500  &   24.33  &   24.65 \\
\hline
\end{tabular}
\end{center}

\begin{figure}[!ht]
    \centering
    \subfloat[RoMie and RoLie Bahavior]{%
      \includegraphics[clip,width=0.9\columnwidth]{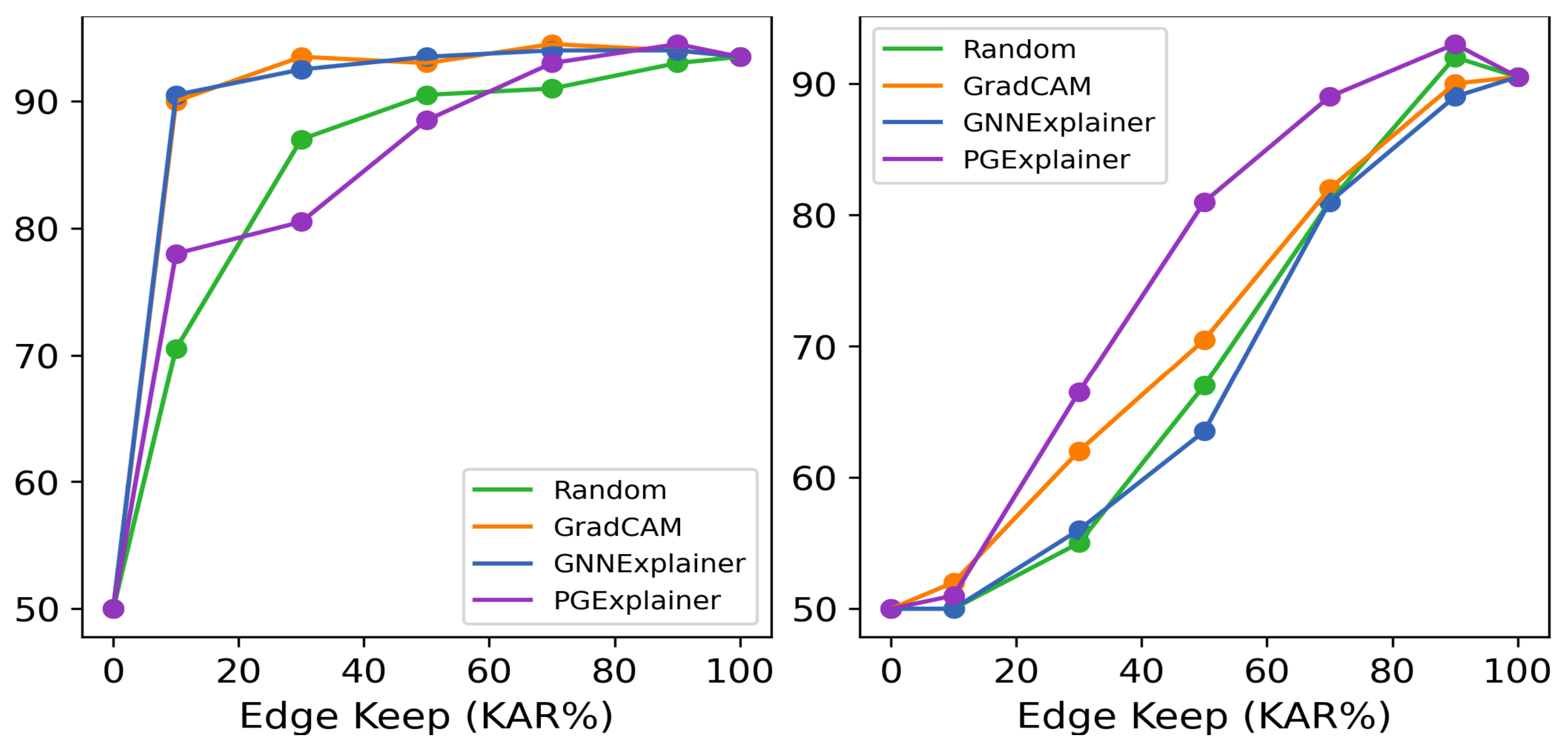}%
    }%
    \caption{\textbf{}
    RoMie performance on unperturbed (up) vs. perturbed (down) evaluation set. Unperturbed version unjustifiably performs better even in the lowest percentages, while informative patterns in samples like discussion-based ones appear in big motifs. on REDDIT-BINARY dataset, using GIN network}
   \label{fig6:RoMieRoLie}
\end{figure}

\begin{figure}[!ht]
    \centering
    \subfloat[RoMie and RoLie Bahavior]{%
      \includegraphics[clip,width=0.45\columnwidth]{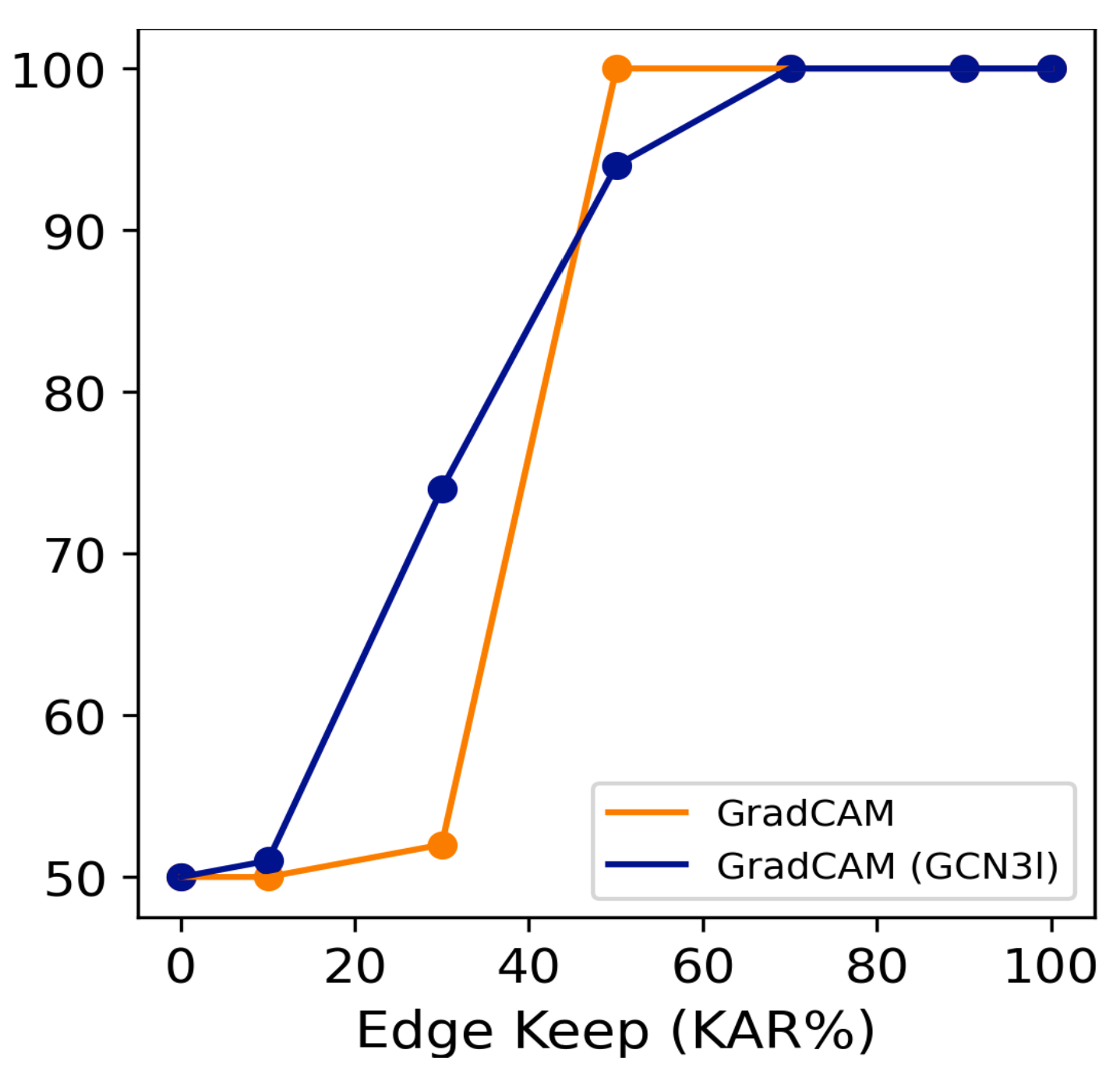}%
    }%
    \caption{\textbf{}
     RoMie evaluation on both GIN and GCN networks and BA-2Motifs dataset using the GradCAM explanation outputs on the GIN network. There is a significant gap in accuracy, particularly in 30\%, meaning GCN uses the features while GIN itself does not. on BA-2Motifs, using GIN an GCN}
   \label{fig6:RoMieRoLie}
\end{figure}

\subsection{Parameters of Attribution Methods}
Alongside random setting, four attributions were used to provide probability edge weighting. For both GNNExplainer and PGExplainer, we applied Adam optimizer with different learning rates of 0.01 and 0.003, and a range number of epochs from 300 to 500 and 10 to 100 were considered to depend on the dataset respectively. For PGExplainer, we have selected a two-layer fully connected network to provide edge weights. For SubgraphX, the configuration is 10 iterations with 14 expansions to extend the child nodes of the tree search. Despite other attributions, SubgraphX does not provide probability weights for edges, therefore we introduced a trick instead in which we have separately generated the binary edge mask for each percentage in our RoMie and RoLie experiments.

\textbf{BA-2Motifs:} A synthetic binary graph classification dataset that each graph possesses either a five-node cycle or house motif [\cite{ref2}] and we expect the attribution to find these patterns. BA-3Motifs: For further examination of our experiments on multi-class datasets, we prepared an extended version of BA-2Motifs where we propose a set of newly generated 500 graphs. We produced the third class by replacing the triangle motif with the house one in the second class. REDDIT-BINARY: A real social networking dataset, each graph belongs to either answer-question or discussion-based threads [\cite{ref1}]. We expect that explainer methods should mostly provide focal-like and sub-group patterns for these two classes respectively.

\textbf{IMDB-BINARY:} It is a real social networking dataset representing movie collaboration between actors/actresses [\cite{ref3}]. Each actor/actress is represented by a node and there is an edge between them if they appeared in the same movie. Each graph is labeled as either action or romance genre, though action has a higher priority.

\begin{figure}[!ht]
    \centering
    \subfloat[RoMie and RoLie Bahavior]{%
      \includegraphics[clip,width=0.9\columnwidth]{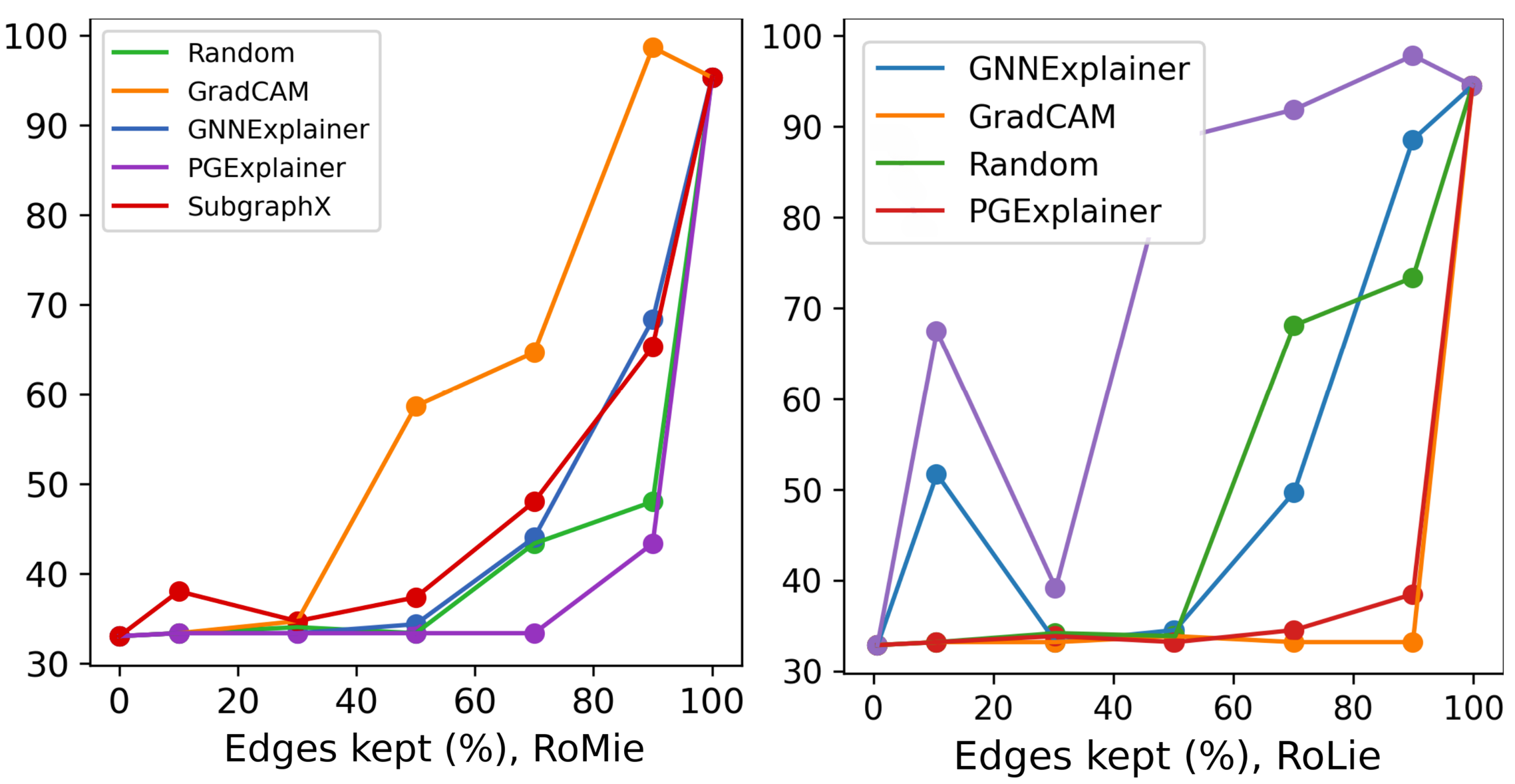}%
    }%
    \\
    \subfloat[Label zero (cycle)]{%
      \includegraphics[clip,width=0.75\columnwidth]{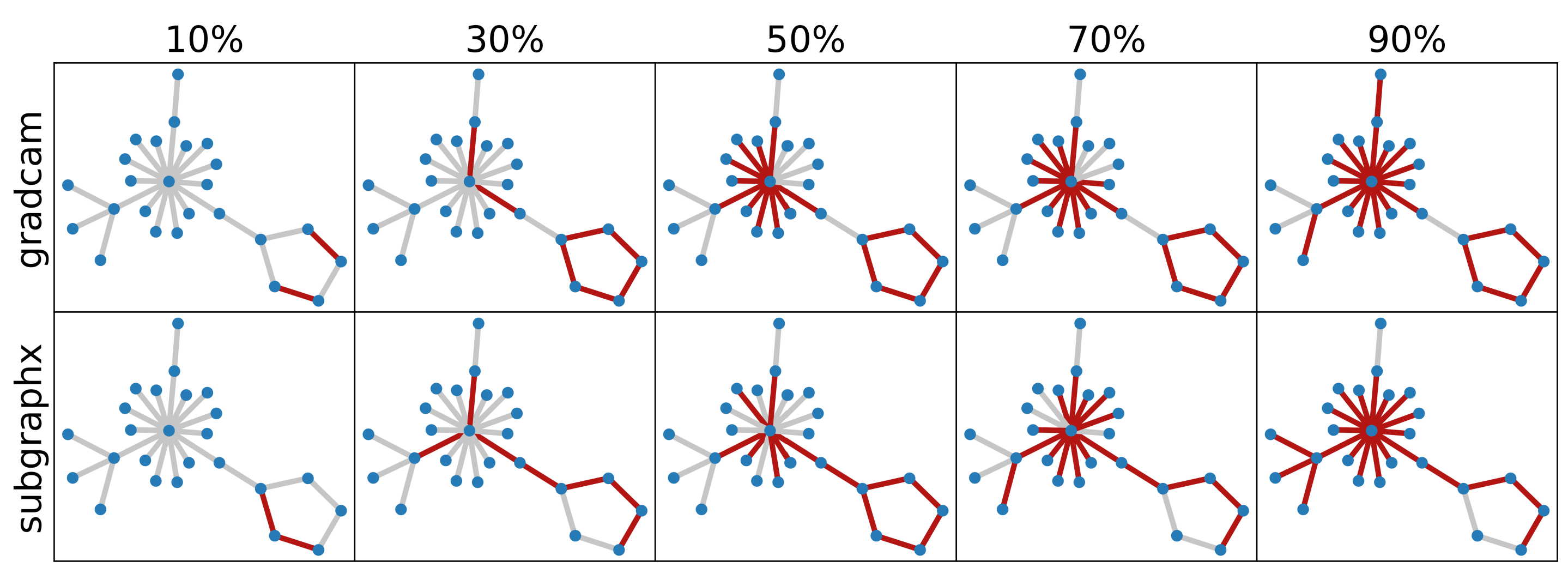}%
    }%
    \\
    \subfloat[Label one (house)]{%
      \includegraphics[clip,width=0.75\columnwidth]{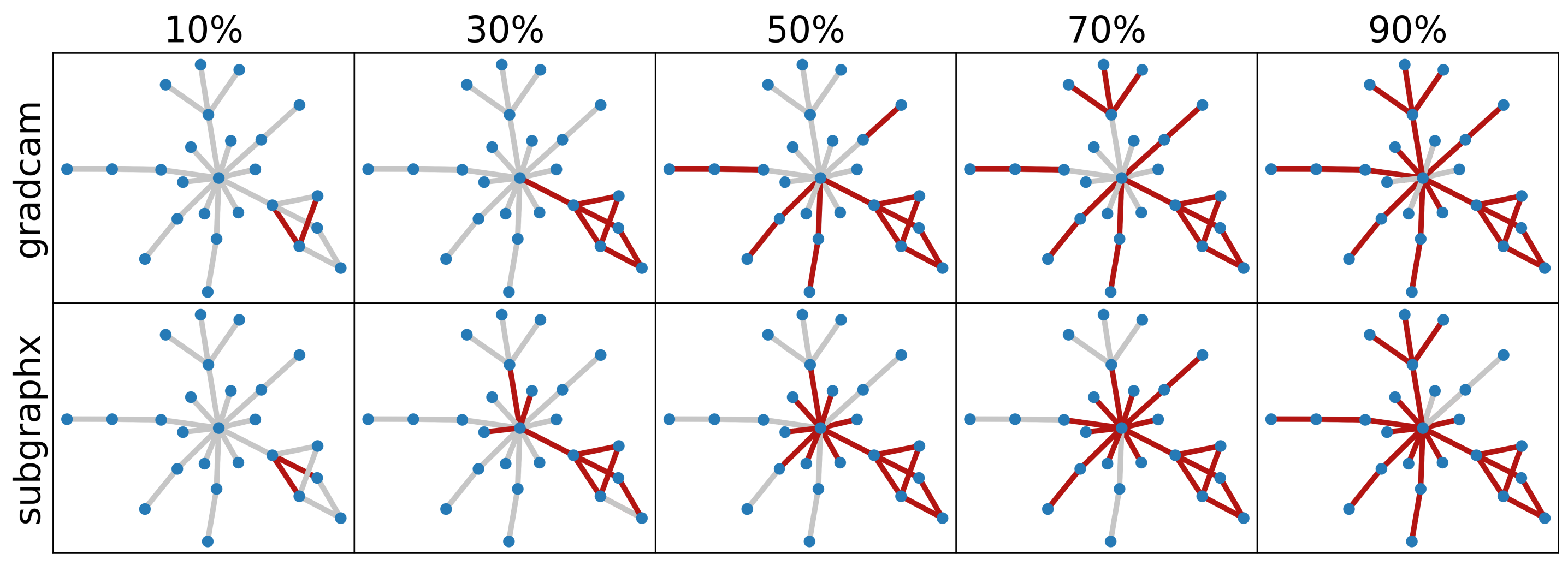}%
    }%
    \\
    \subfloat[Label one (house)]{%
      \includegraphics[clip,width=0.75\columnwidth]{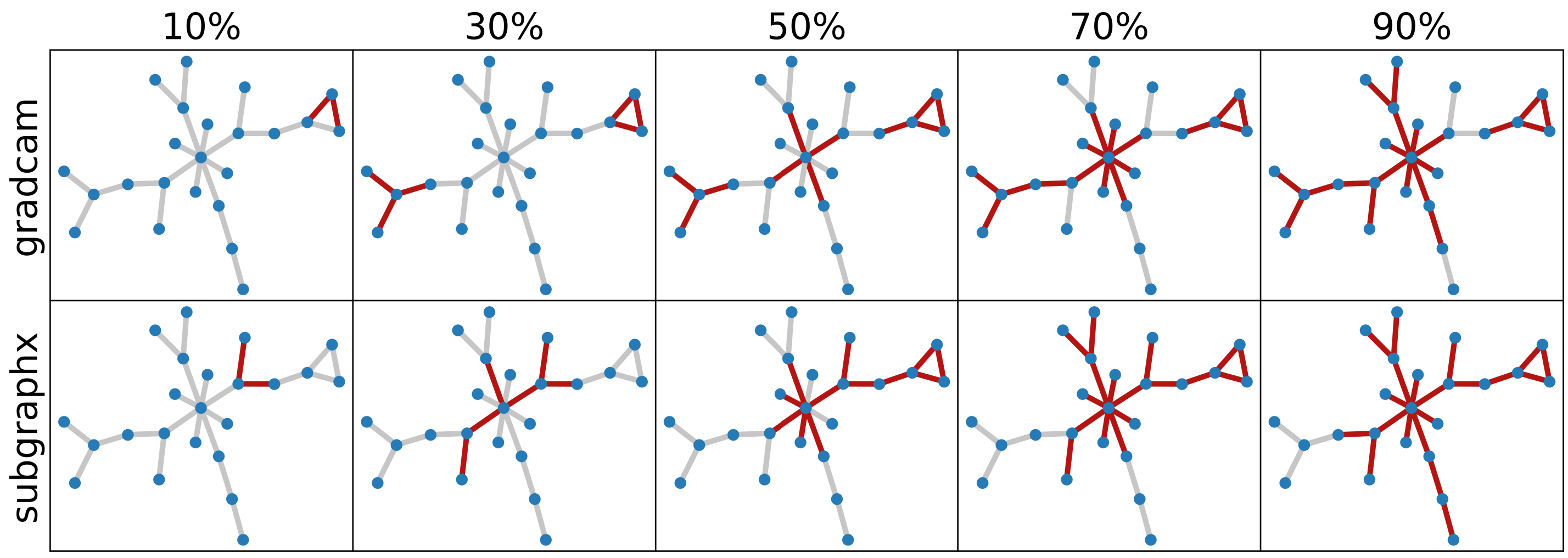}%
    }
    \caption{\textbf{Complementary RoMie and RoLie:} RoMie and RoLie evaluation (a) and outputs of attribition methods (b, c, d). Keeping edges returned by GradCAM and SubgraphX can highly aid the network during retraining while removing them can highly defect the network. Based on the explanation outputs, both methods capture the cycle-house-triangle trio at 50\%, while SubgraphX cannot surpass GradCAM in performance during RoMie. This shows that, unlike GCN, the GIN network replaces other parts of the graph with this trio for its efficient training. on BA3Motifs using GIN network}
   \label{fig6:RoMieRoLie}
\end{figure}

\begin{figure}[!ht]
    \centering
    \subfloat[RoMie and RoLie Bahavior]{%
      \includegraphics[clip,width=0.9\columnwidth]{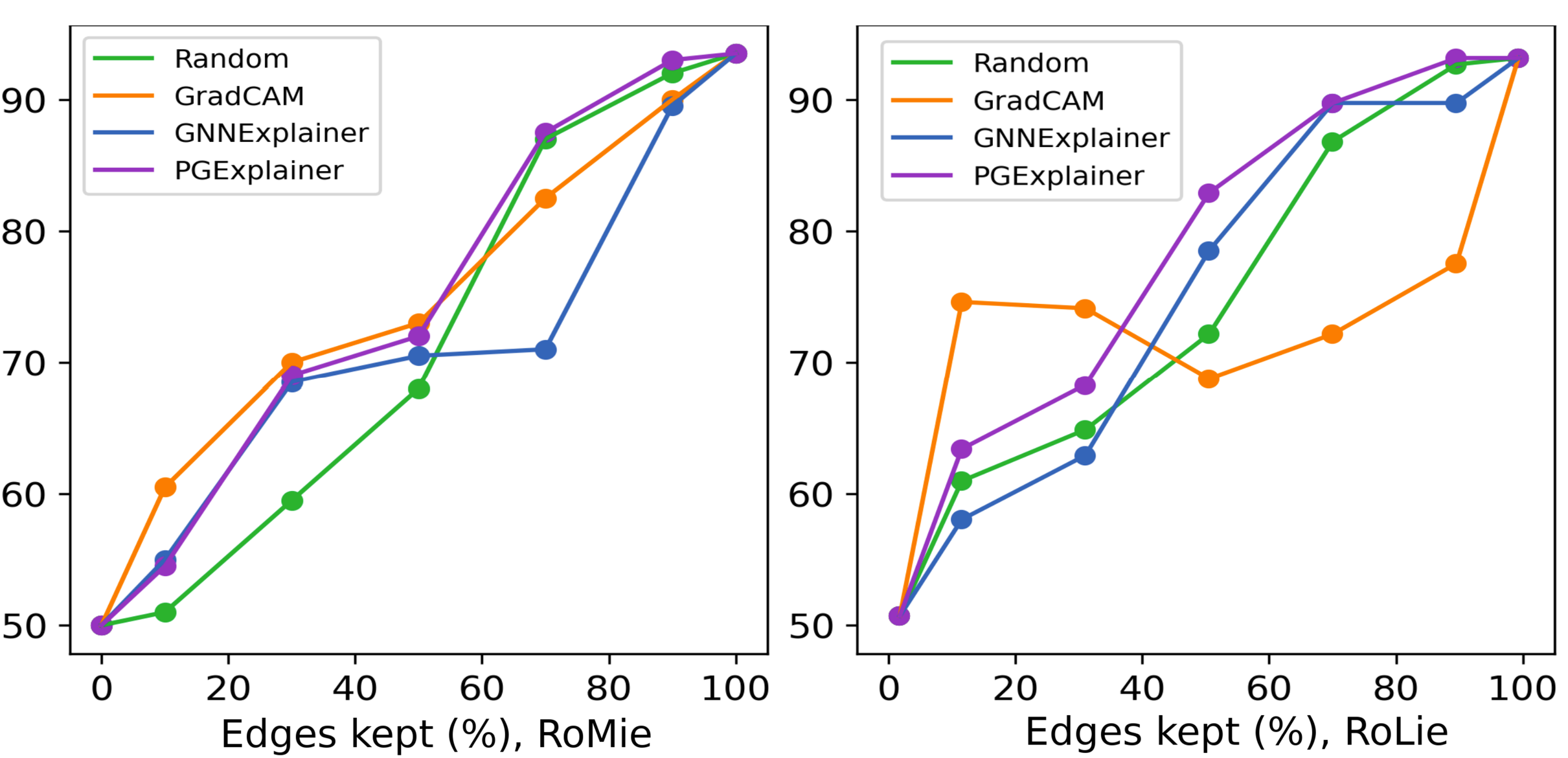}%
    }%
    \\
    \subfloat[Label zero (cycle)]{%
      \includegraphics[clip,width=0.80\columnwidth]{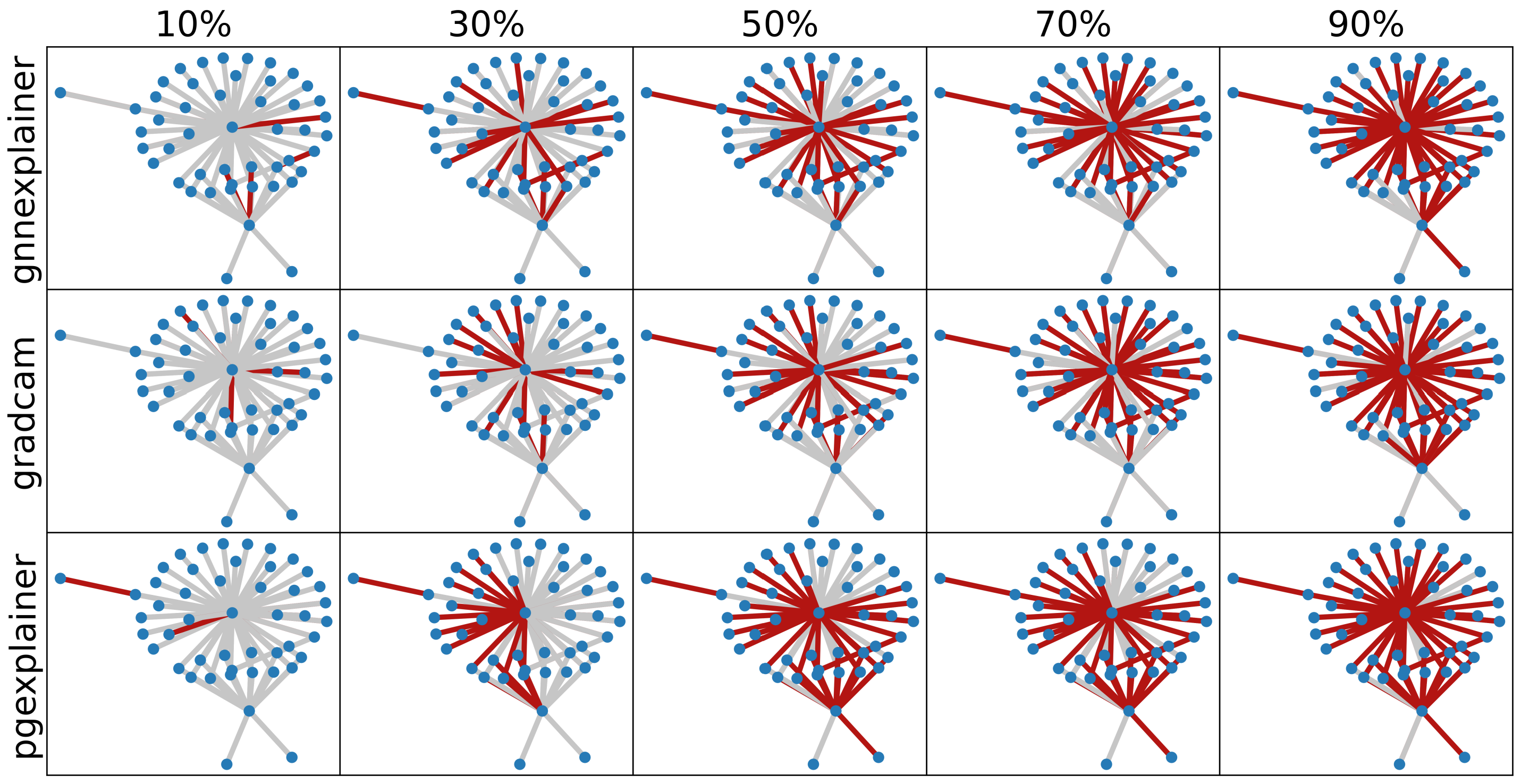}%
    }%
    \\
    \subfloat[Label one (house)]{%
      \includegraphics[clip,width=0.80\columnwidth]{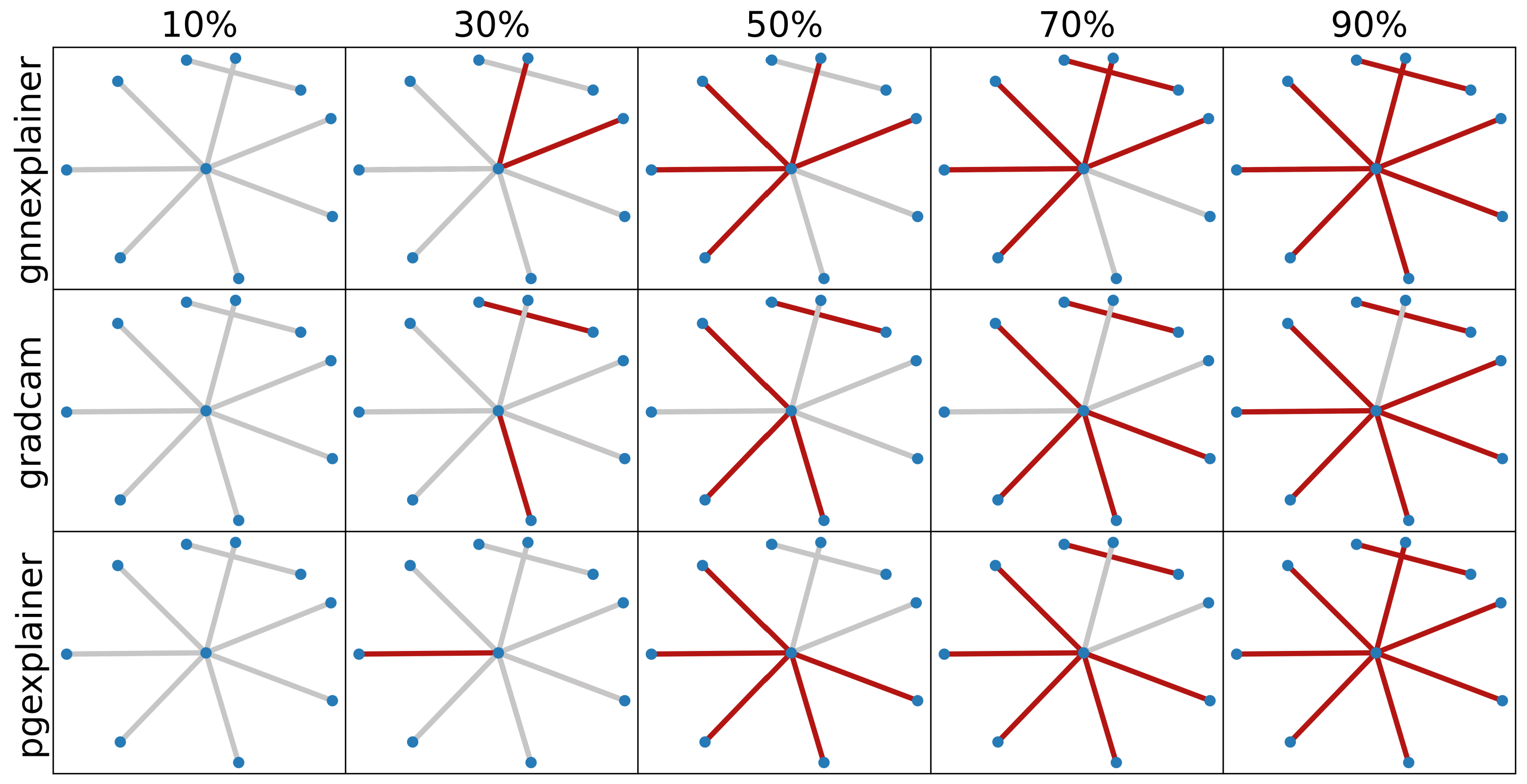}%
    }
    \caption{\textbf{}
    RoMie-RoLie evaluation (a) and outputs of explanation methods (b, c) on REDDIT-BINARY dataset, using a GIN network. Keeping the edges provided by GradCAM can help the network during the retraining than the other explainers, while removing them highly disorders the network performance, particularly between 0-50\%. Based on their explanation outputs, GradCAM returns the focal motifs for question-based samples and multiple sub-group patterns for the discussion-based samples more concretely. Therefore, we can conclude that GIN network focuses on these edges for this dataset.}
   \label{fig6:RoMieRoLie}
\end{figure}

\textbf{MUTAG:} A real biological binary classification dataset determining whether a given chemical compound has a mutagenic effect on a bacterium or not [\cite{ref4}]. Nodes and edges are representing atoms and bonds between them respectively.

\textbf{ENZYME:} It contains enzyme graphs from the BRENDA
enzyme database which is a dataset of protein tertiary structures [\cite{ref5}]. The main goal is to classify each enzyme into
one of the 6 EC top-level classes.


\begin{figure}[!ht]
    \centering
    \subfloat[RoMie and RoLie Bahavior]{%
      \includegraphics[clip,width=0.9\columnwidth]{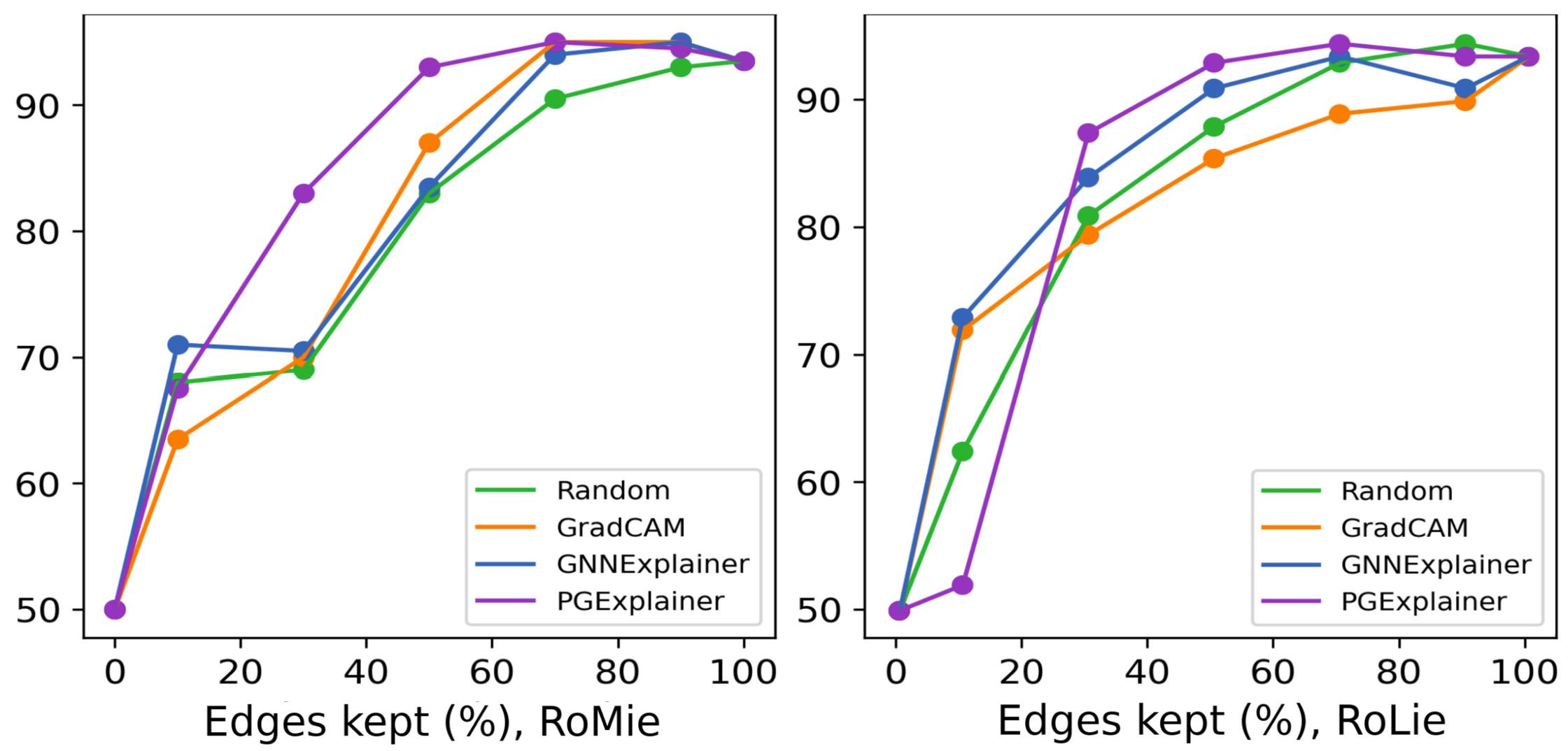}%
    }%
    \caption{\textbf{}RoMie-RoLie evaluation on REDDIT-BINARY dataset using GCN network}
   \label{fig6:RoMieRoLie}
\end{figure}

\begin{figure}[!ht]
    \centering
    \subfloat[RoMie and RoLie Bahavior]{%
      \includegraphics[clip,width=0.9\columnwidth]{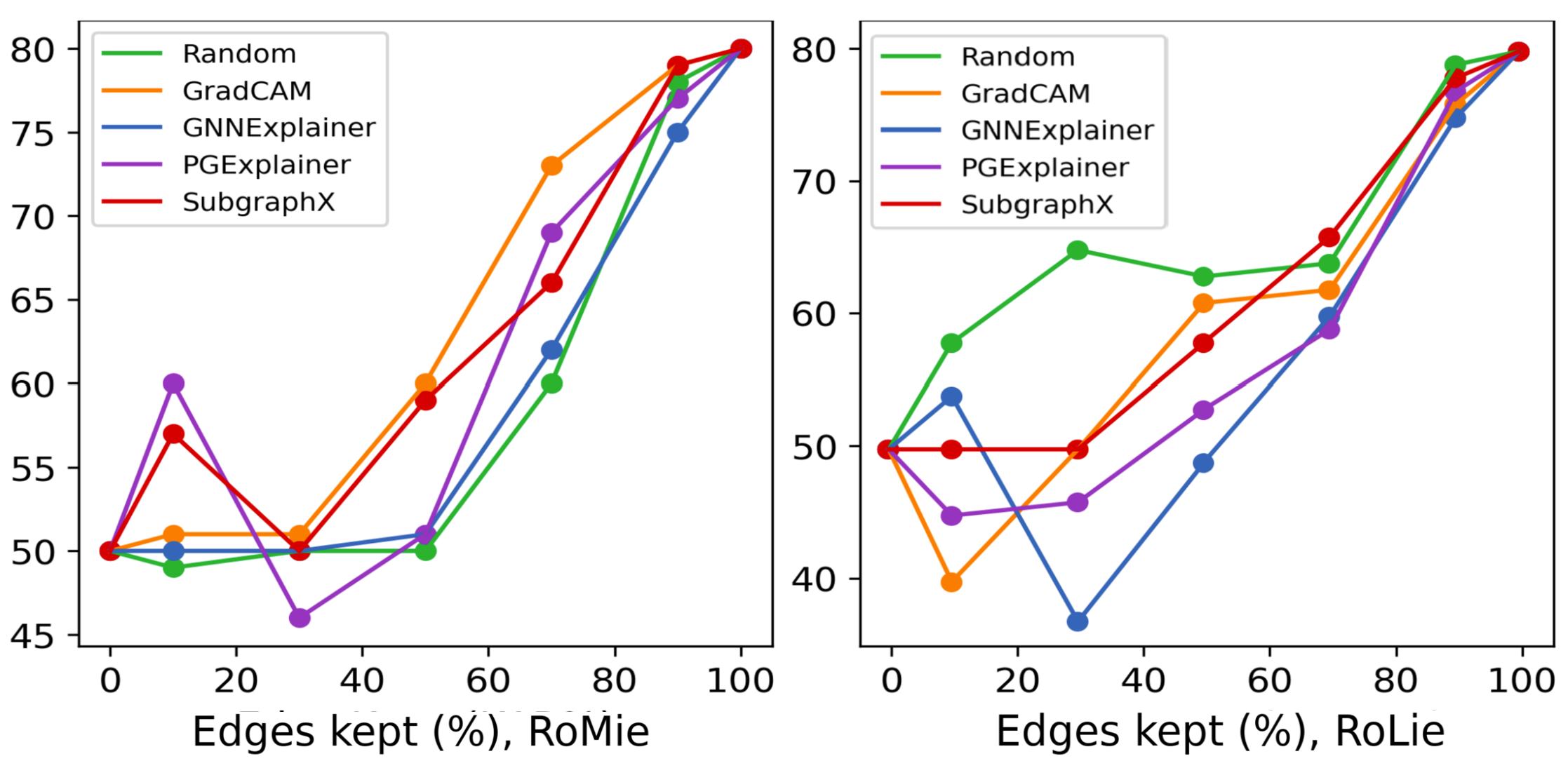}%
    }%
    \caption{\textbf{Complementary RoMie and RoLie:}
    RoMie-RoLiw evaluation on IMDB-BINARY dataset using GIN network.}
   \label{fig6:RoMieRoLie}
\end{figure}

\begin{figure}[!ht]
    \centering
    \subfloat[RoMie and RoLie Bahavior]{%
      \includegraphics[clip,width=0.9\columnwidth]{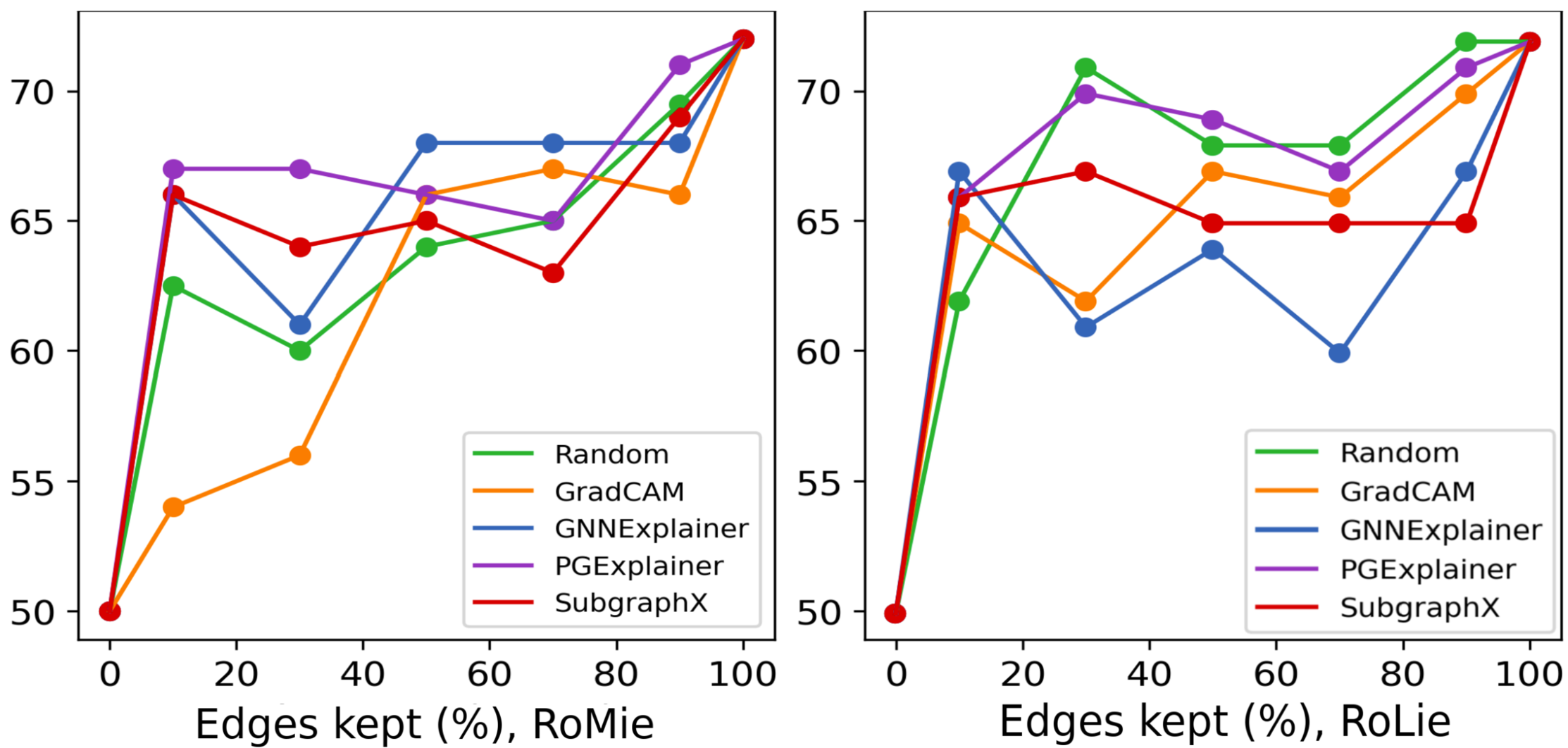}%
    }%
    \caption{\textbf{}
    RoMie-RoLievaluation on IMDB-BINARY dataset using GCN network.}
   \label{fig6:RoMieRoLie}
\end{figure}

\begin{figure}[!ht]
    \centering
    \subfloat[RoMie and RoLie Bahavior]{%
      \includegraphics[clip,width=0.9\columnwidth]{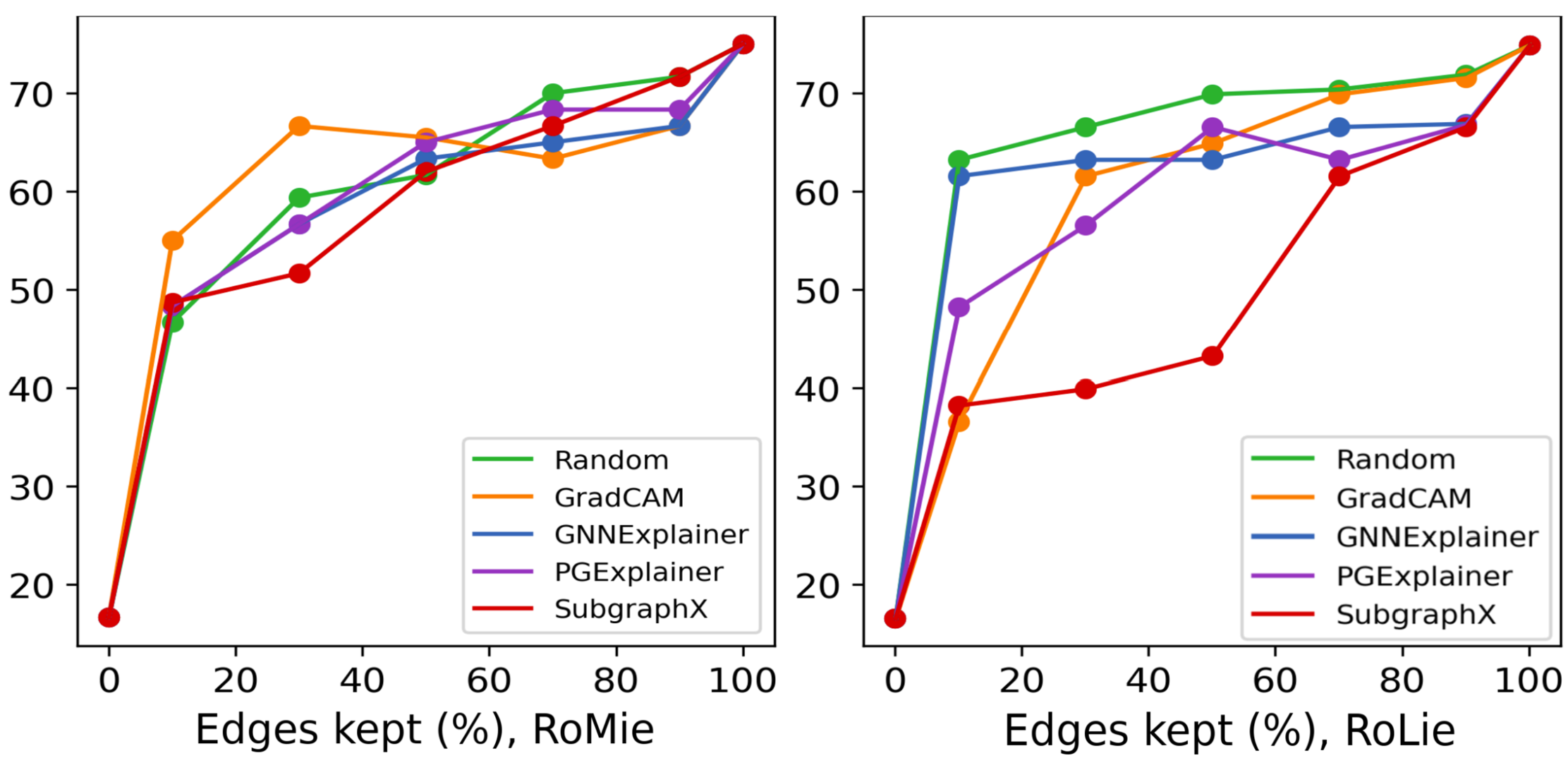}%
    }%
    \caption{\textbf{}
    RoMie-RoLieevaluation on ENZYME dataset using GCN network.}
   \label{fig6:RoMieRoLie}
\end{figure}

\begin{figure}[!ht]
    \centering
    \subfloat[RoMie and RoLie Bahavior]{%
      \includegraphics[clip,width=0.9\columnwidth]{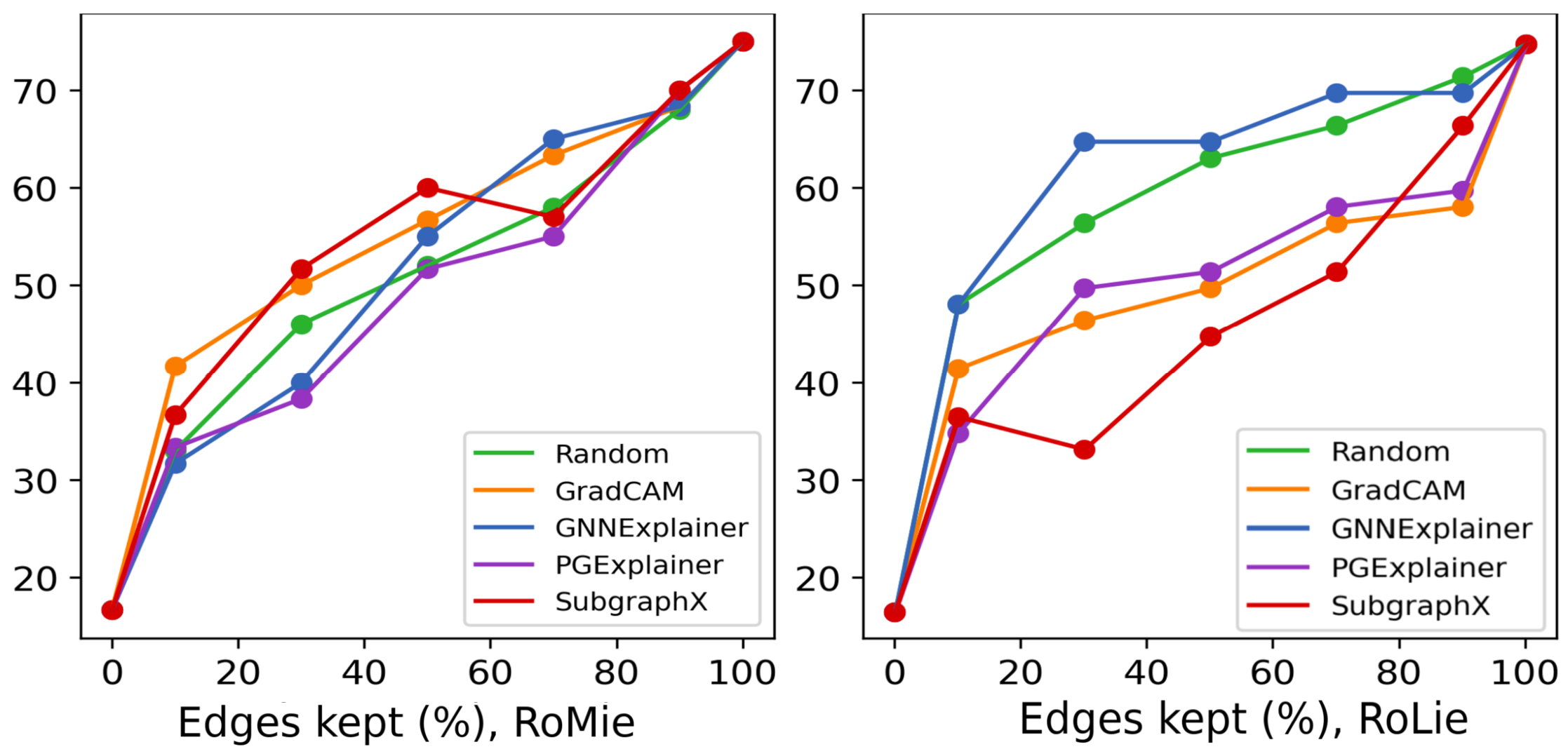}%
    }%
    \caption{\textbf{}
     RoMie-RoLie evaluation on ENZYME dataset using GIN network.}
   \label{fig6:RoMieRoLie}
\end{figure}

\begin{figure}[!ht]
    \centering
    \subfloat[RoMie and RoLie Bahavior]{%
      \includegraphics[clip,width=0.9\columnwidth]{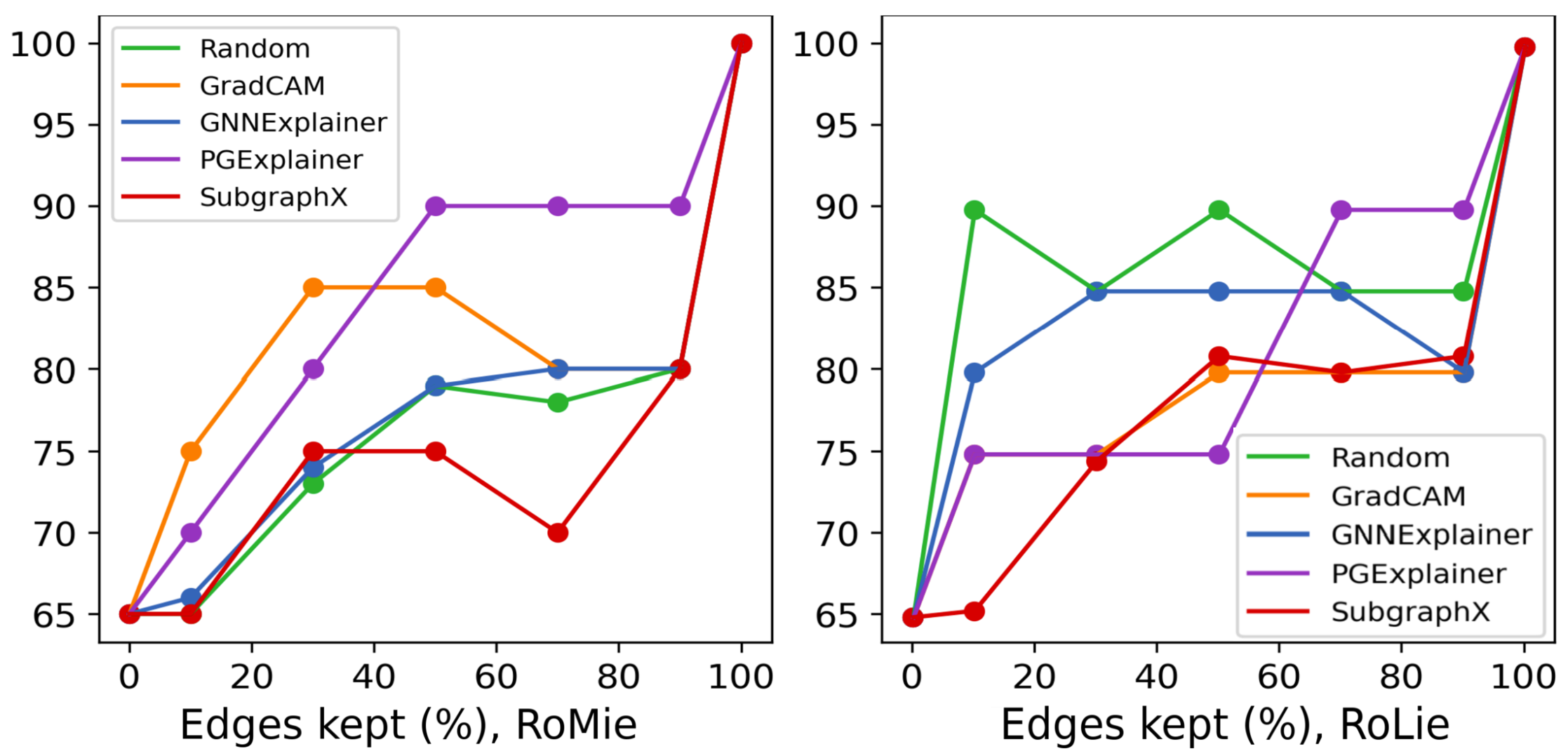}%
    }%
    \caption{\textbf{}
    RoMie-RoLie evaluation on MUTAG dataset using GIN network.}
   \label{fig6:RoMieRoLie}
\end{figure}

\begin{figure}[!ht]
    \centering
    \subfloat[RoMie and RoLie Bahavior]{%
      \includegraphics[clip,width=0.9\columnwidth]{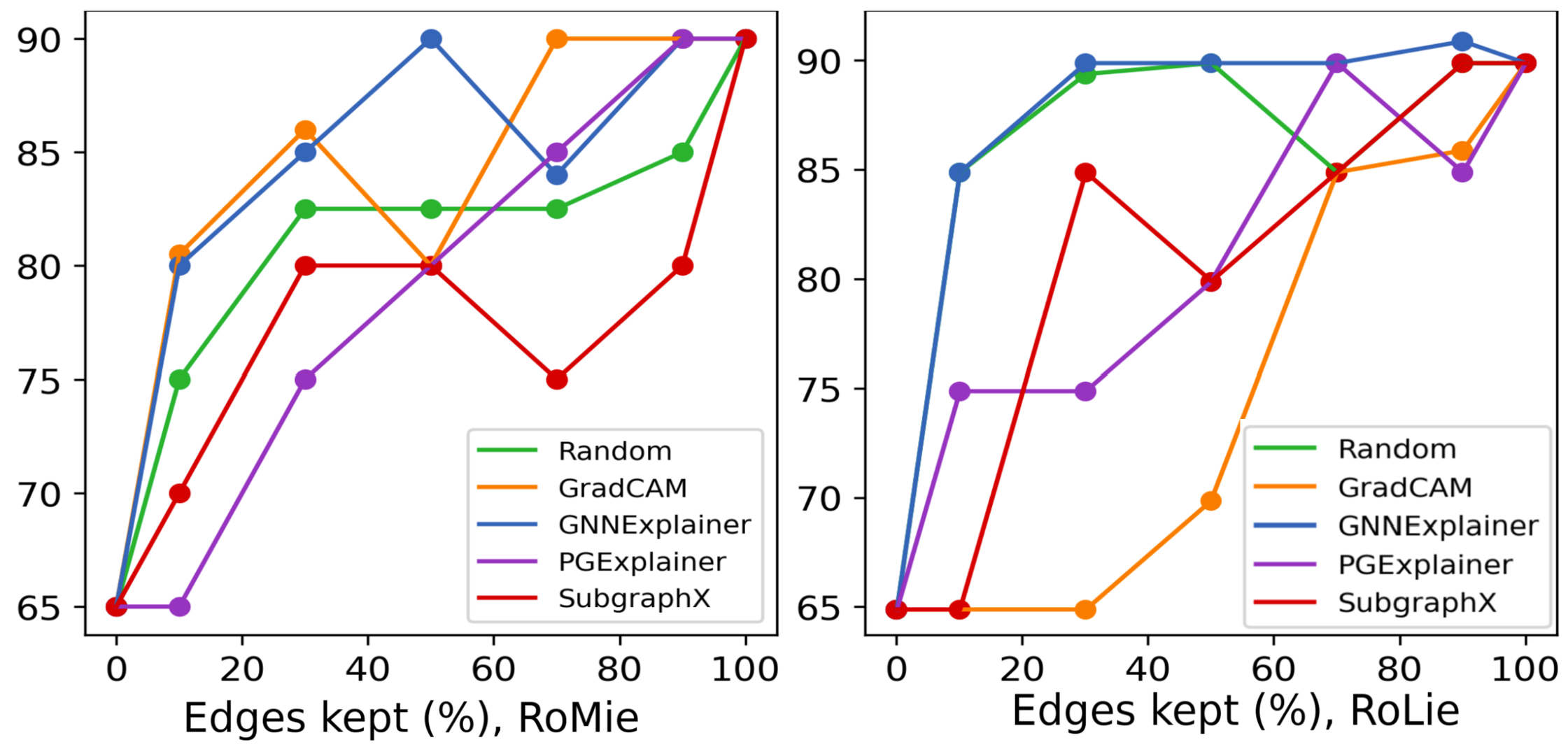}%
    }%
    \caption{\textbf{}
     RoMie-RoLie evaluation on MUTAG dataset using GCN network.}
   \label{fig6:RoMieRoLie}
\end{figure}

\section{Attribution Depend of Datasets and Networks}
Figure 3 represents an evaluation of the GIN network on the BA-3Motifs dataset, where keeping the edges provided by GradCAM and SubgraphX can generally give better discrimination power for the network. By considering their outputs in 70\%, both GradCAM and SubgraphX capture the whole cycle house-triangle trio though SubgraphX could not exceed GradCAM in performance during RoMie. Hence, this means GIN focuses on other features which are not the same as this trio.

Figure 4 represents RoMie-RoLie evaluation on the REDDIT-BINARY dataset using the GIN network, where keeping edges returned by GradCAM can highly assist the network during retraining, and also removing them and keeping the other parts could significantly degrade the re-training performance leading to poor accuracy. Based on the outputs of explanation methods, for question-based samples, GradCAM provides a denser focal motif than the others, which is identical to having a user ask a question and the other ones respond to it; In addition, for the discussion-based samples, GradCAM vividly provides subgroup motifs showing there is a topic and users discuss it in smaller groups with each other. Therefore, we can conclude that the GIN network mostly focuses on the same edges of the input graph proposed by the GradCAM. Due to the numerous nodes per graph on average, SubgraphX analysis takes 40-50X more than other methods, making it impossible to take its output.

\end{document}